\documentclass{article}

\usepackage{microtype}
\usepackage{graphicx}
\usepackage{subcaption}
\usepackage{booktabs} 
\usepackage{siunitx}
\usepackage{longtable}
\usepackage{hyperref}

\usepackage[preprint]{icml2026}

\usepackage{amsmath}
\usepackage{amssymb}
\usepackage{mathtools}
\usepackage{amsthm}
\usepackage{cuted}
\usepackage{multirow}
\usepackage{multicol}
\usepackage{makecell}
\usepackage{algorithm}
\usepackage{algorithmic}
\usepackage{enumitem}
\usepackage{tikz}
\usepackage{pgfplots}
\pgfplotsset{compat=1.18}

\usepackage[capitalize,noabbrev]{cleveref}

\theoremstyle{plain}

\theoremstyle{definition}

\theoremstyle{remark}

\usepackage[textsize=tiny]{todonotes}
\usepackage{xcolor}

\icmltitlerunning{Equivariant Evidential Deep Learning for Interatomic Potentials}

\begin{document}

\twocolumn[
  \icmltitle{Equivariant Evidential Deep Learning for Interatomic Potentials}



  \icmlsetsymbol{equal}{*}

  \begin{icmlauthorlist}
    \icmlauthor{Zhongyao Wang}{equal,fudan,ailab,sii}
    \icmlauthor{Taoyong Cui}{equal,ailab,cuhk}
    \icmlauthor{Jiawen Zou}{fudan}
    \icmlauthor{Shufei Zhang}{ailab}
    \icmlauthor{Bo Yan}{fudan}
    \icmlauthor{Wanli Ouyang}{ailab,cuhk}
    \icmlauthor{Weimin Tan \textsuperscript{$\dagger$}}{fudan}
    \icmlauthor{Mao Su \textsuperscript{$\dagger$}}{ailab,siat}
  \end{icmlauthorlist}

  \icmlaffiliation{ailab}{Shanghai Artificial Intelligence Laboratory, Shanghai, China }
  \icmlaffiliation{fudan}{College of Computer Science and Artificial Intelligence, Fudan University, Shanghai, China}
  \icmlaffiliation{sii}{Shanghai Innovation Institution, Shanghai, China}
  \icmlaffiliation{cuhk}{The Chinese University of Hong Kong,Hong Kong SAR, China}
  \icmlaffiliation{siat}{Shenzhen Institute of Advanced Technology, Chinese Academy of Sciences, Shenzhen, China}

  \icmlcorrespondingauthor{Weimin Tan}{wmtan@fudan.edu.cn}
  \icmlcorrespondingauthor{Mao Su}{sumao@pjlab.org.cn}

  \icmlkeywords{Equivariant Neural Networks, Uncertainty Quantification, Interatomic Potentials, Evidential Deep Learning}

  \vskip 0.3in
]



\printAffiliationsAndNotice{}  

\begin{abstract}
Uncertainty quantification (UQ) is critical for the reliable deployment of machine learning interatomic potentials (MLIPs) in molecular dynamics (MD) simulations.
Evidential deep learning (EDL) offers a theoretically grounded single-model alternative to expensive ensemble methods, estimating both aleatoric and epistemic uncertainty in a single forward pass.
However, extending EDL from scalar targets to vector-valued atomic forces introduces a fundamental challenge: the predicted covariance must transform equivariantly under rotations to maintain statistical self-consistency.
We propose $\text{e}^2$IP, a backbone-agnostic evidential framework that represents force uncertainty as a full $3\times3$ symmetric positive definite (SPD) covariance tensor, constructed via a tangent-space parameterization and Riemannian exponential map that jointly guarantees positive definiteness and $SO(3)$ equivariance.
Experiments on liquid water, MD22, rMD17, and silica glass benchmarks show that $\text{e}^2$IP achieves better calibration (CE, ES, NLL) than deep ensembles --- including heteroskedastic variants --- while retaining single-model inference efficiency with ${\sim}4.5\times$ speedup.
\end{abstract}

\section{Introduction}
\label{sec:intro}

Machine-learning interatomic potentials (MLIPs) enable large-scale molecular dynamics (MD) by approximating \textit{ab initio} potential energy surfaces at dramatically reduced cost~\cite{behler2007generalized}. 
With recent advances in $E(3)$-equivariant neural architectures, MLIPs can now achieve near first-principles accuracy for energies and atomic forces, substantially extending the accessible time- and length-scales of atomistic simulation and supporting a broad range of applications from materials property prediction to non-equilibrium processes~\cite{batzner2022nequip,batatia2022mace,thomas2018tfn,gasteigergemnet,wood2025family,musaelian2023learning}. 
As a result, MLIPs are becoming a key tool for bridging first-principles modeling and large-scale atomistic simulation~\cite{jia2020pushing,kozinsky2023scaling}.

Despite these advances, deploying MLIPs reliably in long-timescale MD remains challenging~\cite{perez2025uncertainty}. 
Force errors can become problematic when simulations enter out-of-distribution configurations, where ML force fields may produce pathological energy or force predictions~\cite{fuforces}.
Reliable and computationally efficient uncertainty quantification (UQ) is therefore essential for safe deployment~\cite{bilbrey2025uncertainty}, as well as for uncertainty-aware workflows such as data selection and adaptive model refinement.

Ensembles are widely used and often regarded as a robust ``gold standard'' for UQ in MLIPs~\cite{gal2016dropout,lakshminarayanan2017simple,wollschlager2023uncertainty,willow2025bayesian}, but their training and inference costs scale linearly with the number of models, making them expensive for large-scale simulation.
Moreover, recent work has suggested that scaling a single model can be equally effective as full ensembling in certain settings, and that ensemble calibration may require careful post-hoc adjustment~\cite{rahaman2021uncertainty,abe2022deep,dern2024theoretical}.
These limitations motivate single-model UQ approaches.
Evidential deep learning (EDL) is a promising direction: by predicting parameters of a prior distribution, it estimates both aleatoric and epistemic uncertainty in one forward pass~\cite{sensoy2018evidential,amini2020deep,deng2023uncertainty}.
However, most EDL methods target scalar properties~\cite{soleimany2021evidential}.
Xu \textit{et al.}~proposed eIP~\cite{xu2025evidential} for forces, but model uncertainty as independent per-axis variances (diagonal covariance in a fixed global frame).
This cannot capture cross-axis correlations and, more critically, cannot transform consistently under rigid rotations --- breaking the coherence of the probabilistic model across orientations.

This limitation is not merely aesthetic.
Forces are equivariant vectors ($\mathbf{F}\mapsto R\mathbf{F}$), so by standard transformation rules the covariance must satisfy $\Sigma' = R\Sigma R^\top$.
Equivariance of the uncertainty tensor is therefore not an additional physical assumption but a \emph{requirement for distributional self-consistency}.

\begin{figure*}[htbp]
  \centering
    \centering
    \includegraphics[width=\linewidth]{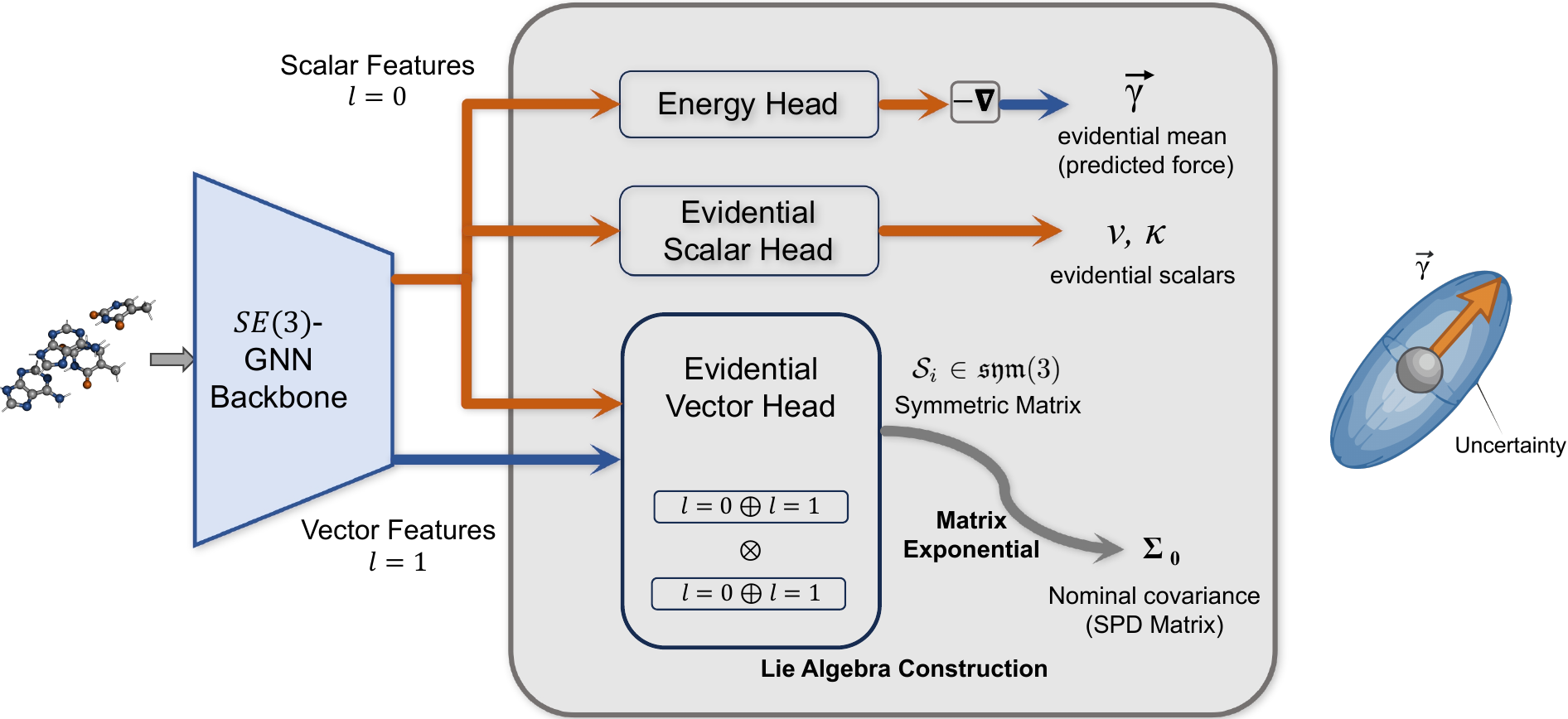}
    \caption{
\textbf{e$^{2}$IP framework}.
An SE(3)-equivariant GNN predicts the force mean and NIW evidential parameters, and constructs an equivariant SPD covariance $\Sigma_0$ (via a tangent-space parameterization and Riemannian exponential map) to produce tensor-valued uncertainty in a single forward pass.
}
    \label{fig:overview}
\end{figure*}

We propose $\text{e}^2$IP (Fig.~\ref{fig:overview}), an equivariant evidential framework that represents force uncertainty as a full $3\times3$ SPD covariance tensor.
Positive definiteness and $SO(3)$ equivariance are jointly guaranteed by parameterizing a symmetric matrix in the tangent space of $\mathbb{S}_{++}^3$ and mapping it via the Riemannian exponential~\cite{moakher2005differential,arsigny2007geometric}.
Built on a Normal--Inverse--Wishart (NIW) evidential model, $\text{e}^2$IP decomposes uncertainty into aleatoric and epistemic components in a single forward pass.
An equivariance-preserving spectral stabilizer prevents the numerical degeneracy that otherwise causes training to diverge.
Experiments on four molecular benchmarks show that $\text{e}^2$IP achieves better calibration and ranking quality than deep ensembles --- including heteroskedastic (MVE) and shallow variants --- at ${\sim}4.5\times$ lower inference cost.

\paragraph{Contributions.}
\begin{itemize}
    \item We introduce $\text{e}^{2}$IP, an $SO(3)$-equivariant evidential framework that models aleatoric and epistemic uncertainty in vector-valued forces as full $3\times3$ SPD covariances in a single forward pass.

    \item We propose an $SO(3)$-consistent full-covariance construction via a tangent-space parameterization and Riemannian exponential map on $\mathbb{S}_{++}^3$, together with equivariance-preserving spectral stabilization for numerically robust training.

    \item Through extensive validation across diverse molecular benchmarks, $\text{e}^{2}$IP demonstrates better uncertainty calibration and data efficiency at the single-model inference cost.
\end{itemize}

\section{Preliminaries}
\label{sec:preliminaries}

This section establishes the notation and the two key requirements---equivariance and positive definiteness---that motivate our covariance construction.

\paragraph{Problem setup.}
We consider an atomic configuration $\mathcal{G}$ with coordinates
$\{\vec x_i \in \mathbb{R}^3\}_{i=1}^N$ and species $\{z_i \in \mathbb{Z}\}_{i=1}^N$.
The goal is to predict atom-wise forces
$\{\vec y_i \in \mathbb{R}^3\}_{i=1}^N$ together with a per-atom $3\times3$ covariance matrix representing the uncertainty of each force prediction.

\paragraph{Equivariant backbone.}
We assume an $SE(3)$-equivariant graph neural network $f_\theta$ (e.g., NequIP~\cite{batzner2022nequip}, MACE~\cite{batatia2022mace})
that maps $\mathcal{G}$ to node-wise features organized by angular momentum order $l$:
for each atom $i$, the backbone outputs features
$\{\mathbf{h}_i^{l}\}$ including scalars ($l=0$) and higher-order tensors ($l>0$), each transforming as an irreducible representation (irrep) of $SO(3)$.
Task-specific heads then map these features to the force mean and uncertainty parameters.

\subsection{Equivariance requirements for mean and covariance}
\label{subsec:equiv_cov}

Let $g=(R,\mathbf{t})\in SE(3)$ with $R\in SO(3)$.
Forces are equivariant vectors: rotating a molecular configuration rotates the forces accordingly, so the predicted force mean $\vec\gamma_i\in\mathbb{R}^3$ (the model's best estimate of the force on atom $i$) must satisfy
\begin{equation}
\label{eq:mean_equiv}
\vec\gamma_i(g\cdot \mathcal{G}) = R\,\vec\gamma_i(\mathcal{G}).
\end{equation}
\emph{The same requirement extends to uncertainty.} If the force uncertainty is represented by a covariance matrix
$\mathbf{\Sigma}_i \in \mathbb{S}^3_{++}$ (the set of $3\times3$ symmetric positive definite matrices), it must transform as a rank-2 tensor:
\begin{equation}
\label{eq:cov_equiv}
\mathbf{\Sigma}_i(g\cdot \mathcal{G}) = R\,\mathbf{\Sigma}_i(\mathcal{G})\,R^\top .
\end{equation}
Predicting independent per-axis variances (a diagonal matrix in a fixed global basis)
violates Eq.~\eqref{eq:cov_equiv} under non-trivial rotations and cannot capture anisotropic
correlations aligned with local geometric directions. This is the central limitation of existing diagonal evidential approaches~\cite{xu2025evidential}.

\subsection{Irrep decomposition of symmetric matrices}
\label{subsec:cg_decomp}

To construct equivariant symmetric matrices from backbone features, we leverage $SO(3)$ representation theory.
Any $\mathbf{S}_i\in\mathrm{sym}(3)$ decomposes into an isotropic (trace) and a traceless part:
\begin{equation}
\label{eq:sym_decomp}
\mathbf{S}_i
=
\underbrace{\tfrac{1}{3}\mathrm{tr}(\mathbf{S}_i)\,\mathbf{I}}_{\text{scalar }(0e)}
\;+\;
\underbrace{\left(\mathbf{S}_i - \tfrac{1}{3}\mathrm{tr}(\mathbf{S}_i)\,\mathbf{I}\right)}_{\text{traceless }(2e)}.
\end{equation}
The scalar ($0e$) channel captures the isotropic scale (1 degree of freedom), and the
five-dimensional traceless ($2e$) channel captures the anisotropic shape.
This $(0e\oplus 2e)$ structure provides a natural, rotation-compatible parameterization for constructing
$\mathbf{S}_i$ from equivariant features, which we exploit in Sec.~\ref{subsec:parameterization}.

\section{Methodology}
\label{sec:method}

\paragraph{Overview.}
We propose \textit{Equivariant Evidential Deep Learning for Interatomic Potentials} ($\text{e}^2$IP), a single forward pass
probabilistic framework for atom-wise vector regression with rotation-consistent uncertainty (Fig.~\ref{fig:overview}).
Given an input structure, an $SE(3)$-equivariant network jointly predicts the force mean and a \emph{full} $3\times3$
SPD uncertainty tensor for each atom.
Uncertainty is modeled using a Normal--Inverse--Wishart (NIW) evidential formulation~\cite{meinert2021multivariate}, yielding a multivariate
Student-$t$ predictive distribution with explicit aleatoric and epistemic decomposition.
The method builds on existing components --- SE(3)-equivariant GNN backbones, CG tensor products for coupling irreps, and multivariate evidential regression via NIW --- and introduces two novel contributions:
(i) to our knowledge, the first full $3\times3$ equivariant SPD covariance parameterization for evidential force uncertainty, achieved via a tangent-space parameterization and Riemannian exponential map that jointly guarantees positive definiteness, SO(3) equivariance, and stable NIW training; and
(ii) an equivariance-preserving spectral stabilizer via coefficient-space damping that bounds eigenvalue magnitude while preserving equivariance, without which training diverges (Fig.~\ref{fig:condnum_ablation}).
\paragraph{Notation.}
We set $d=3$. For atom $i$: $\vec y_i\in\mathbb{R}^d$ is the target force vector; $\vec\gamma_i\in\mathbb{R}^d$ is the predicted force mean; $\vec\mu_i$ and $\mathbf{\Sigma}_i$ are latent mean and covariance (random variables under the NIW prior); $\mathbf{\Sigma}_{0,i}\in\mathbb{S}_{++}^3$ is the network-predicted nominal covariance (deterministic); $\nu_i>0$ and $\kappa_i>0$ are evidence parameters controlling covariance and mean uncertainty, respectively. $SE(3)$ denotes the special Euclidean group; $\mathbb{S}_{++}^3$ the manifold of $3\times3$ symmetric positive definite matrices; $\mathrm{sym}(3)$ the vector space of $3\times3$ real symmetric matrices; and ``irrep'' stands for irreducible representation of $SO(3)$. A comprehensive notation table is provided in Appendix~\ref{app:notation}.

\subsection{Multivariate Evidential Learning Framework}
\label{subsec:generative_process}

Having established that force uncertainty requires an equivariant SPD covariance (Sec.~\ref{subsec:equiv_cov}), we now describe the probabilistic model that jointly represents force predictions and their uncertainty, then show how it decomposes into aleatoric and epistemic components.

\paragraph{Generative model.}
For each atom $i$, we model the observed force $\vec{y}_i$ as drawn from a Gaussian with unknown mean and covariance:
\begin{equation}
    p(\vec{y}_i \mid \vec{\mu}_i, \mathbf{\Sigma}_i) = \mathcal{N}(\vec{y}_i \mid \vec{\mu}_i, \mathbf{\Sigma}_i).
\end{equation}
Rather than estimating $\vec{\mu}_i$ and $\mathbf{\Sigma}_i$ directly, we treat them as latent random variables and place a conjugate Normal--Inverse--Wishart (NIW) prior~\cite{meinert2021multivariate}:
\begin{equation}
\label{eq:niw_prior_main}
\mathbf{\Sigma}_i \sim \mathcal{W}^{-1}(\mathbf{\Psi}_i,\nu_i),\qquad
\vec{\mu}_i \mid \mathbf{\Sigma}_i \sim \mathcal{N}\!\left(\vec{\gamma}_i, \frac{1}{\kappa_i}\mathbf{\Sigma}_i\right).
\end{equation}
Here the network predicts four quantities per atom: the force mean $\vec{\gamma}_i$, a nominal covariance $\mathbf{\Sigma}_{0,i}$ (a deterministic matrix that controls the shape of $\mathbf{\Sigma}_i$), and two scalar evidence parameters $\nu_i$ (covariance evidence) and $\kappa_i$ (mean evidence).
We reparameterize the IW scale matrix as
\begin{equation}
\label{eq:psi_reparam_main}
\mathbf{\Psi}_i \triangleq \nu_i \mathbf{\Sigma}_{0,i},
\end{equation}
so that $\mathbf{\Sigma}_{0,i}\in\mathbb{S}_{++}^3$ directly describes the \emph{shape} of the aleatoric noise.

\paragraph{Posterior predictive.}
Marginalizing the latent $(\vec{\mu}_i,\mathbf{\Sigma}_i)$ yields a closed-form multivariate Student-$t$ predictive distribution (derivation in Appendix~\ref{app:student_t_nll}):
\begin{equation}
\label{eq:student_t_main}
\begin{split}
    p(\vec{y}_i \mid \vec{\gamma}_i, \mathbf{\Sigma}_{0,i}, \nu_i, \kappa_i)
= \mathrm{St}_{m_i}\!\left(\vec{y}_i \,\middle|\, \vec{\gamma}_i,\; \mathbf{\Lambda}_i \right),
\\
m_i \triangleq \nu_i-d+1,
\quad
\mathbf{\Lambda}_i \triangleq \frac{\nu_i(\kappa_i + 1)}{\kappa_i (\nu_i - d + 1)} \mathbf{\Sigma}_{0,i}.
\end{split}
\end{equation}
The degrees of freedom $m_i$ control the heaviness of the tails: low $m_i$ (little evidence) produces heavy tails reflecting high uncertainty, while $m_i\to\infty$ recovers a Gaussian.

\paragraph{Aleatoric and epistemic uncertainty decomposition.}
The NIW hierarchy induces an uncertainty decomposition:
\begin{equation}
\begin{split}
    \label{eq:ale_epi_main}
&\mathbf{U}_{\text{ale}} \triangleq \mathbb{E}[\mathbf{\Sigma}_i]
= \frac{\nu_i \mathbf{\Sigma}_{0,i}}{\nu_i - d - 1}
\\
&\mathbf{U}_{\text{epi}} \triangleq \mathrm{Var}[\vec{\mu}_i]
= \frac{1}{\kappa_i}\,\mathbb{E}[\mathbf{\Sigma}_i]
= \frac{\nu_i \mathbf{\Sigma}_{0,i}}{\kappa_i(\nu_i - d - 1)}
\end{split}
\end{equation}
$\mathbf U_{\text{ale}}$ reflects irreducible observation noise (e.g., thermal fluctuations or DFT numerical noise), while $\mathbf U_{\text{epi}}$ captures uncertainty in the predicted mean force due to insufficient or unrepresentative training data. Both are full $3\times3$ SPD tensors with physical directional meaning: for example, $\mathbf{U}_{\text{ale}}$ may be anisotropic along a soft vibrational mode. Crucially, because $\mathbf{\Sigma}_{0,i}$ is constructed equivariantly (Sec.~\ref{subsec:parameterization}), both $\mathbf{U}_{\text{ale}}$ and $\mathbf{U}_{\text{epi}}$ transform correctly under $SO(3)$ --- rotating the molecule rotates the uncertainty ellipsoid accordingly.

\paragraph{Constraints and stable scalar parameterization.}
We require $\kappa_i>0$ and $\nu_i>d+1$ so that $\mathbb{E}[\mathbf{\Sigma}_i]$ exists in Eq.~\eqref{eq:ale_epi_main}.
In practice we enforce a stronger lower bound $\nu_i \ge d+2$ to avoid the near-singular regime
of $\nu_i/(\nu_i-d-1)$. We enforce constraints via shifted softplus:
\begin{equation}
\begin{split}
    &\nu_i = \mathrm{softplus}(\hat{\nu}_i) + (d+2)\\
    &\kappa_i = \mathrm{softplus}(\hat{\kappa}_i) + \epsilon
\end{split}
\label{eq:scalar_constraints_main}
\end{equation}
where $\epsilon=10^{-6}$.

\subsection{Equivariant SPD covariance via tangent-space parameterization}
\label{subsec:parameterization}

The evidential framework above requires the nominal covariance $\mathbf{\Sigma}_{0,i}$ to be simultaneously (a) symmetric positive definite, (b) equivariant under $SO(3)$, and (c) parameterized in a way amenable to gradient-based optimization. Standard parameterizations such as Cholesky decomposition satisfy (a) and (c) but not (b): the triangular factor $L$ in $\mathbf{\Sigma}_{0,i}=LL^\top$ does not transform predictably under rotation. We address all three requirements with a tangent-space construction.

\paragraph{From tangent space to SPD via the Riemannian exponential map.}
The manifold $\mathbb{S}_{++}^3$ is a Riemannian symmetric space whose tangent space at the identity is $\mathrm{sym}(3)$~\cite{moakher2005differential,arsigny2007geometric}. The matrix exponential $\exp\colon\mathrm{sym}(3)\to\mathbb{S}_{++}^3$ is the Riemannian exponential map --- a diffeomorphism that maps any unconstrained symmetric matrix to a valid SPD matrix.
We predict $(0e\oplus 2e)$ irrep coefficients from the equivariant backbone, map them to a symmetric matrix
$\mathcal{S}_i\in\mathrm{sym}(3)$ via a fixed Clebsch--Gordan basis (Sec.~\ref{subsec:cg_decomp}), and define
\begin{equation}
\label{eq:matrix_exp_main}
    \mathbf{\Sigma}_{0,i} = \exp(\mathcal{S}_i).
\end{equation}
Because $\mathcal{S}_i$ is symmetric, it admits an eigen-decomposition $\mathcal{S}_i=U\Lambda U^\top$ with real
eigenvalues; hence $\exp(\mathcal{S}_i)=U\exp(\Lambda)U^\top$ has strictly positive eigenvalues and is therefore SPD.
Moreover, the matrix exponential is conjugation-equivariant: for any $R\in SO(3)$,
\begin{equation}
\label{eq:exp_equiv_main}
    \exp(R\mathcal{S}_iR^\top) = R\,\exp(\mathcal{S}_i)\,R^\top,
\end{equation}
which follows from the power-series definition of $\exp(\cdot)$ and the identity $(R\mathcal{S}_iR^\top)^k
= R\mathcal{S}_i^kR^\top$. Consequently, $\mathbf{\Sigma}_{0,i}$ transforms equivariantly as a rank-2 tensor.~\cite{shumaylovlie}

\paragraph{Irrep-to-Cartesian map.}
Let $\mathbf{z}_i=[s_i,\mathbf{t}_i]\in\mathbb{R}^{1+5}$ denote the predicted coefficients, where $s_i$ is the $l=0$
channel and $\mathbf{t}_i$ parameterizes the $l=2$ (symmetric traceless) channel. We use a fixed linear map to obtain $\mathcal{S}_i\in\mathrm{sym}(3)$; details (including the constant map
$\mathbf{Q}$ and vectorization convention) are provided in Appendix~\ref{app:Q_map}.

\subsection{Spectral stabilization via coefficient-space damping}
\label{subsec:damping}

The construction above guarantees that $\mathbf{\Sigma}_{0,i}$ is always SPD and equivariant. However, in practice the matrix exponential amplifies extreme eigenvalues: if the network predicts large-magnitude $\mathcal{S}_i$, the resulting $\mathbf{\Sigma}_{0,i}$ becomes severely ill-conditioned (eigenvalue ratios $>10^5$), causing numerical failures in the Cholesky factorization required by the NLL loss (Fig.~\ref{fig:condnum_ablation}, right). To prevent this without breaking equivariance or requiring explicit eigendecomposition, we damp the irrep coefficients before constructing $\mathcal{S}_i$.

\paragraph{Equivariance-preserving damping.}
For the scalar channel, we apply a smooth two-sided damper $\tilde{s}_i=\phi_{\mathrm{low}}(\phi_{\mathrm{high}}(s_i))$.
For the $l=2$ channel, we damp only the magnitude:
\begin{equation}
\label{eq:l2_damp_main}
\tilde{\mathbf{t}}_i = \alpha_i\,\mathbf{t}_i,
\qquad
\alpha_i \triangleq \frac{\phi_{\mathrm{high}}(\|\mathbf{t}_i\|_2)}{\|\mathbf{t}_i\|_2 + \epsilon}.
\end{equation}
Since $\|\mathbf{t}_i\|_2$ is rotation-invariant, $\alpha_i$ is invariant, and the operation preserves equivariance by
construction. We then form $\tilde{\mathbf{z}}_i=[\tilde{s}_i,\tilde{\mathbf{t}}_i]$, map it to $\mathcal{S}_i$,
and compute $\mathbf{\Sigma}_{0,i}=\exp(\mathcal{S}_i)$. The explicit piecewise linear–Tanh form of
$\phi_{\mathrm{high}}/\phi_{\mathrm{low}}$ is provided in Appendix~\ref{app:damping_details}.

\subsection{Training objective}
\label{subsec:loss}

We now describe how the full pipeline --- backbone $\to$ evidential heads $\to$ tangent-space construction $\to$ spectral damping $\to$ matrix exponential --- is trained end-to-end. Algorithm~\ref{alg:forward} summarizes the forward pass.
Given labeled data $\{(\mathcal{G}_j, \{\vec{y}_i^j\})\}$ where $\mathcal{G}_j$ is an atomic graph and $\vec{y}_i^j\in\mathbb{R}^3$ is the DFT force on atom $i$, the network outputs $(\vec{\gamma}_i, \mathbf{\Sigma}_{0,i}, \nu_i, \kappa_i)$ per atom. The loss has two components.

\paragraph{Student-$t$ negative log-likelihood.}
Minimizing the negative log-likelihood of Eq.~\eqref{eq:student_t_main} yields the training loss in Eq.~\eqref{eq:loss}. Intuitively, the $\log|\mathbf{\Sigma}_{0,i}|$ term penalizes overconfident covariance (too small $\mathbf{\Sigma}_{0,i}$ is costly), while the Mahalanobis term $\text{M}$ penalizes force prediction errors weighted by the predicted uncertainty.
For completeness, the expanded form used in the implementation is reported in Appendix~\ref{app:student_t_nll}.
\begin{equation}
\begin{split}
\mathcal{L}_{\mathrm{NLL},i}
&= \log\Gamma\!\left(\frac{\nu_i-d+1}{2}\right)
 - \log\Gamma\!\left(\frac{\nu_i+1}{2}\right)\\
 & \quad+ \frac{d}{2}\log\!\left(\frac{\pi\nu_i(1+\kappa_i)}{\kappa_i}\right)
 + \frac{1}{2}\log\left|\mathbf{\Sigma}_{0,i}\right| \\
&\quad
 + \frac{\nu_i+1}{2}\log\!\left(
1 + \frac{\kappa_i}{\nu_i(1+\kappa_i)}\text{M}
\right).\\
\\
\text{M} &\triangleq  (\vec{y}_i-\vec{\gamma}_i)^\top \mathbf{\Sigma}_{0,i}^{-1}(\vec{y}_i-\vec{\gamma}_i)
\end{split}
\label{eq:loss}
\end{equation}

\paragraph{Evidence regularization.}
To discourage overconfident evidence on hard or out-of-distribution samples, we add a monotonic penalty on evidence:
\begin{equation}
\label{eq:reg_loss_main}
    \mathcal{L}_{\text{reg,i}} = (\nu_i + \kappa_i) \|\vec{y}_i - \vec{\gamma}_i\|_2,
\end{equation}
and optimize $\mathcal{L}_{\text{total}} = \mathcal{L}_{\text{NLL}} + \lambda \mathcal{L}_{\text{reg}}$.
Note that $\partial \mathcal{L}_{\text{reg}}/\partial \nu_i = \|\vec{y}_i - \vec{\gamma}_i\|_2$, so larger errors induce
a stronger gradient opposing evidence accumulation.

\paragraph{Scalar proxy for tensor epistemic uncertainty.}
For comparison against scalar error metrics, we report
\begin{equation}
\label{eq:scalar_proxy_main}
    u_{\text{scalar}} = \sqrt{\mathrm{tr}(\mathbf{U}_{\text{epi}})/3}.
\end{equation}
The trace is rotation-invariant ($\mathrm{tr}(R\mathbf{A}R^\top)=\mathrm{tr}(\mathbf{A})$), providing a consistent
scalar summary.

\begin{algorithm}[t]
\caption{Forward pass and loss computation for atom $i$. Steps 1--3 use the shared backbone; steps 4--9 are the evidential uncertainty head; steps 10--14 compute the loss.}
\label{alg:forward}
\begin{algorithmic}[1]
\STATE $\{\mathbf h_i^\ell\} \leftarrow f_\theta(G)$, \COMMENT{SE(3)-equivariant backbone}
\STATE $\boldsymbol E \leftarrow \text{EnergyHead}(\{\mathbf h_i^0\})$, \COMMENT{energy}
\STATE $\boldsymbol\gamma_i \leftarrow -\nabla \boldsymbol E $, \COMMENT{vector mean}
\STATE $(\hat\nu_i,\hat\kappa_i) \leftarrow \text{MLP}(\mathbf h_i^{0})$
\STATE $\nu_i \leftarrow \mathrm{softplus}(\hat\nu_i) + (d+2)$; $\kappa_i \leftarrow \mathrm{softplus}(\hat\kappa_i)+\epsilon$
\STATE $(s_i,\mathbf t_i) \leftarrow \text{EquivariantHead}(\{\mathbf h_i^\ell\})$,\COMMENT{$0e\oplus 2e$ coeffs}
\STATE $(\tilde s_i,\tilde{\mathbf t}_i) \leftarrow \text{Damp}(s_i,\mathbf t_i)$
\STATE $S_i \leftarrow \text{SymMap}([\tilde s_i,\tilde{\mathbf t}_i]) \in \mathrm{sym}(3)$
\STATE $\Sigma_{0,i} \leftarrow \exp(S_i)$
\STATE Compute $\mathcal{L}_{\mathrm{NLL},i}$
\STATE Compute $\mathcal{L}_{\mathrm{reg},i}$
\STATE Compute $\mathcal{L}_{\mathrm{energy}}$
\STATE $\mathcal{L}_{i,\mathrm{forces}} \gets \mathcal{L}_{\mathrm{NLL},i} + \lambda_{\mathrm{reg}} \mathcal{L}_{\mathrm{reg},i}$
\STATE \textbf{Return} $\mathcal{L}_i \gets \lambda_{\mathrm{energy}}\mathcal{L}_{i,\mathrm{energy}}+\lambda_{\mathrm{forces}}\mathcal{L}_{i,\mathrm{forces}}$
\end{algorithmic}
\end{algorithm}

\section{Experiments}
\label{sec:experiments}

We evaluate $\text{e}^{2}$IP across five experimental settings designed to probe different aspects of uncertainty quality:
\begin{itemize}[nosep,leftmargin=*]
    \item \textbf{Sec.~\ref{subsec:water_experiment}}: Probabilistic reliability on \textit{ab initio} liquid water, comparing against deep ensembles (MAP and heteroskedastic), shallow ensembles, and eIP.
    \item \textbf{Sec.~\ref{subsec:data_efficiency}}: Sample efficiency and rotation consistency on MD22 supramolecular systems.
    \item \textbf{Sec.~\ref{subsec:silica_ood}}: Backbone migration and OOD generalization on the \textsc{Silica} glass benchmark.
    \item \textbf{Sec.~\ref{subsec:rmd17}}: Standard force-field evaluation on rMD17 molecular benchmarks.
    \item \textbf{Sec.~\ref{subsec:ablation_damping}}: Ablation of spectral stabilization.
\end{itemize}
Unless otherwise stated, we use AlphaNet~\cite{yin2025alphanet} as the $SE(3)$-equivariant backbone. Evaluation metrics (ES, CE, NLL, Coverage, Spearman) are defined in Appendix~\ref{app:metrics}.

\subsection{Evaluation on \textit{Ab Initio} Liquid Water Thermodynamics}
\label{subsec:water_experiment}

We evaluate on the \textit{ab initio} liquid water benchmark~\cite{cheng2019ab} (revPBE0-D3), a demanding condensed-phase setting where strong anharmonic fluctuations produce configuration-dependent force variability.
This stress-tests whether UQ remains reliable under realistic physical noise --- a prerequisite for uncertainty-aware workflows such as active learning and risk control.

We use an 80\%/10\%/10\% train/validation/test split and report mean $\pm$ std over five random seeds.
We compare against several baselines: (1) a deep ensemble (DE) consisting of five independently MAP-trained models, reported both raw (DE(raw)) and with post-hoc isotropic calibration (DE(cal), $\sigma^2$ fitted on validation set; Appendix~\ref{app:ensemble_sigma_cal}); (2) a heteroskedastic deep ensemble (MVE-DE) where each of K=5 members predicts both force mean and per-component diagonal variance, trained with Gaussian NLL loss following the standard MVE formulation~\cite{scalia2020evaluating,kellner2024uncertainty}; (3) a shallow ensemble (SE) with shared backbone and K=5 independent output heads~\cite{kellner2024uncertainty}; and (4) the diagonal evidential baseline eIP~\cite{xu2025evidential}.
\begin{figure}
  \centering

  \begin{subfigure}{0.24\textwidth}
    \centering
    \includegraphics[width=\linewidth]{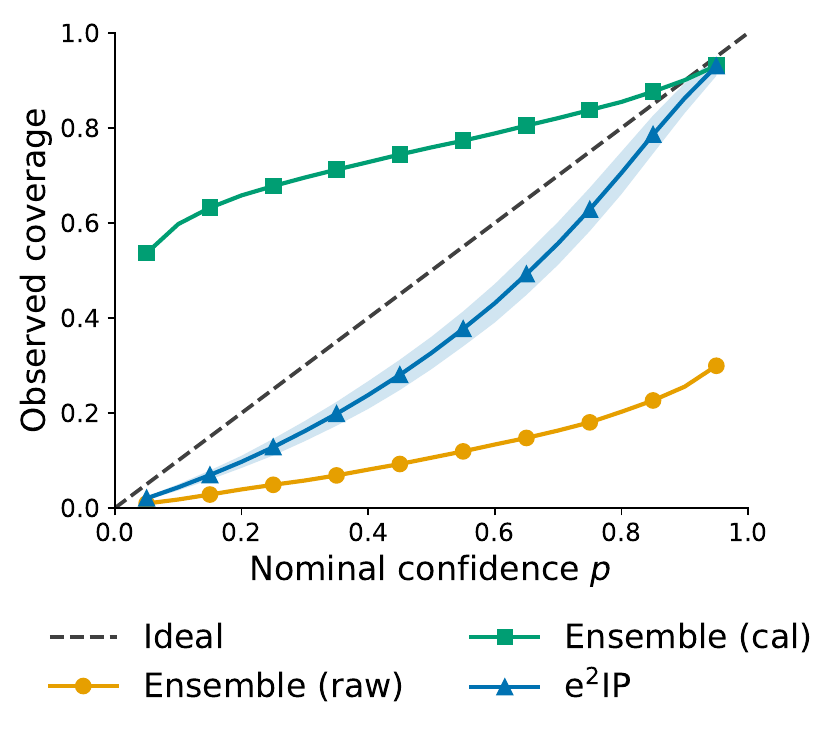}
  \end{subfigure}\hfill
  \begin{subfigure}{0.24\textwidth}
    \centering
    \includegraphics[width=\linewidth]{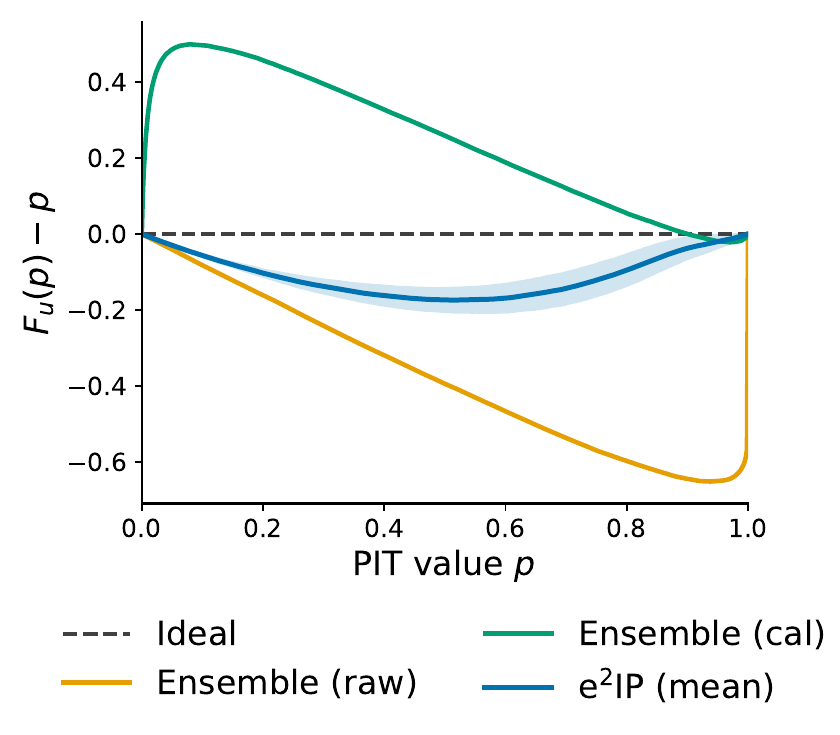}
  \end{subfigure}

  \vspace{0.6em}

  \begin{subfigure}{0.24\textwidth}
    \centering
    \includegraphics[width=\linewidth]{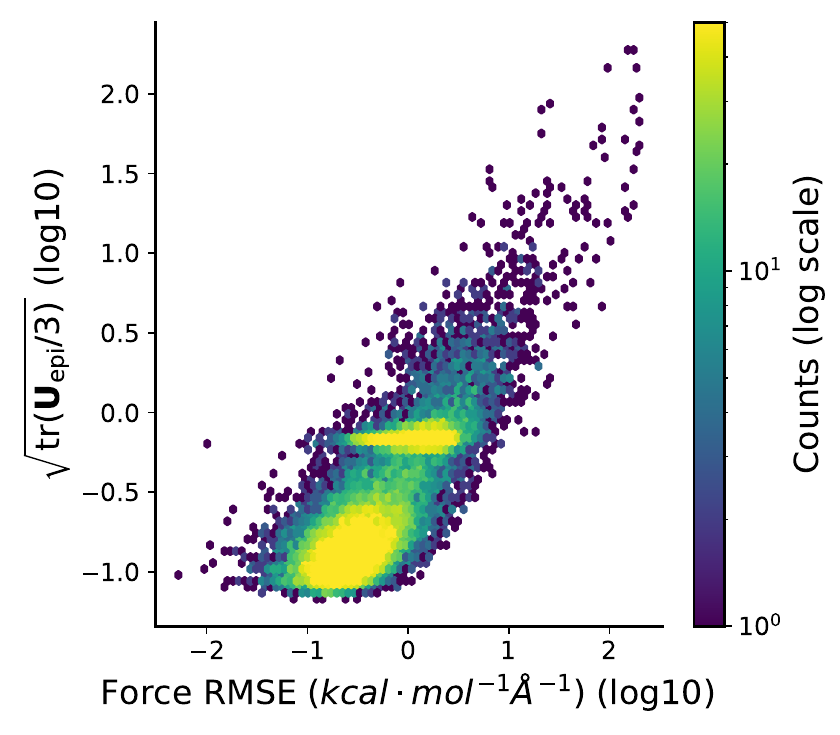}
  \end{subfigure}\hfill
  \begin{subfigure}{0.24\textwidth}
    \centering
    \includegraphics[width=\linewidth]{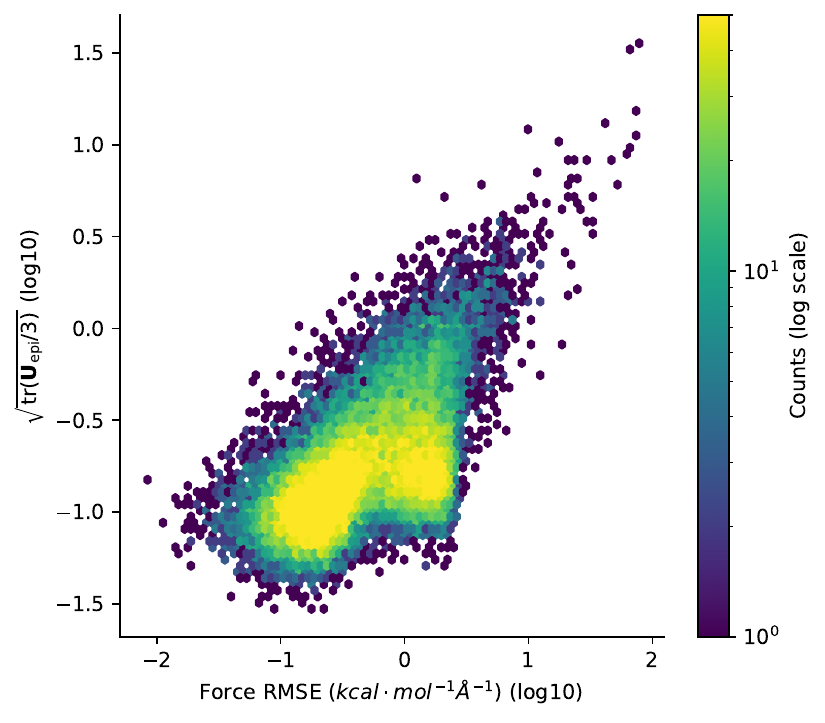}
  \end{subfigure}

  \caption{\textbf{Uncertainty diagnostics on liquid water.}
\emph{Top-left:} reliability diagram (observed vs.\ nominal coverage; dashed line is ideal). 
\emph{Top-right:} PIT CDF deviation $\hat{F}_U(p)-p$ (zero is ideal).
Ensemble (raw) is strongly overconfident, while post-hoc calibration over-corrects and becomes under-confident; $\text{e}^2$IP stays closest to the ideal in both views.
\emph{Bottom:} alignment between force error and predicted epistemic scale, plotting Force RMSE against $\sqrt{\mathrm{tr}(\mathbf{U}_{\text{epi}})/3}$ (both in log10), shown as 2D density (counts in log scale) for $\text{e}^2$IP (left) and Ensemble (right).
$\text{e}^2$IP exhibits a clearer monotonic relationship and higher Spearman correlation (0.78 vs.\ 0.61), indicating better error-aware ranking.}
\label{fig:four}
\end{figure}

Table~\ref{tab:water_results} reports point accuracy (Force MAE), probabilistic quality (Energy Score ES~\cite{gneiting2007strictly}, NLL), and calibration diagnostics (Coverage@95, Calibration Error CE, PIT; Fig.~\ref{fig:four}).
The results show a clear accuracy--efficiency--reliability trade-off: DE achieves the lowest MAE via model averaging but incurs ${\sim}4.5\times$ higher inference cost, while $\text{e}^2$IP attains competitive accuracy in a single forward pass.

\begin{table*}
    \centering
    \caption{
\textbf{Results on the ab initio liquid water benchmark~\citep{cheng2019ab}}.
We report Force MAE (lower is better), Energy Score (ES, lower is better),
Coverage@95 (closer to $0.95$ is better), Calibration Error (CE, lower is better),
Negative log-likelihood (NLL, lower is better), and inference time per atom (ms; lower is better).
All numbers are mean $\pm$ std over five random seeds.
For the ensemble baseline, the isotropic calibration parameter $\sigma^2$ is
fit on the validation set and then fixed for test evaluation. \textbf{Best results are highlighted in bold.}
}
\label{tab:water_results}
    \setlength{\tabcolsep}{4pt}
    \begin{tabular}{lccccccc}
        \toprule
        \textbf{Method} & \makecell{\textbf{Avg. MAE } \\ (\textbf{\unit{kcal/mol/\AA}})}$\downarrow$&\makecell{\textbf{Energy}\\ \textbf{ Score}}$\downarrow$& \makecell{\textbf{Coverage@95} \\ $\approx 0.95$} & \makecell{\textbf{Calibration}\\ \textbf{Error}} $\downarrow$ &\textbf{NLL} $\downarrow$ & \makecell{\textbf{Inference Time}\\ \textbf{per-atom (\unit{ms})}} $\downarrow$\\
        \midrule
        \textbf{DE(raw)}  & \textbf{0.386 $\pm$ 0.012}   & \textbf{0.64 $\pm$ 0.00 }& 0.255 &0.380&237.62 &0.463 $\pm$ 0.067\\
        \textbf{DE(cal)}  & \textbf{0.386 $\pm$ 0.012}   & 0.67 $\pm$ 0.00 & \textbf{0.931} &0.256&2.74 &0.463 $\pm$ 0.067\\
        \textbf{MVE-DE}  & 0.492  & 0.73 & 0.970 &0.139&1.59 &0.463 $\pm$ 0.067\\
        \textbf{SE(cal)} & 0.517 $\pm$ 0.012 & 0.93 $\pm$ 0.06 & 0.790 $\pm$ 0.071 & 0.134 $\pm$ 0.028 & 15.2 $\pm$ 5.6 & 0.234 $\pm$ 0.004 \\
        \textbf{eIP}~\cite{xu2025evidential} & 0.499 $\pm$ 0.016 & 2.15 $\pm$ 2.38 & 0.652 $\pm$ 0.008 & 0.314 $\pm$ 0.008 & 3.13 $\pm$ 0.19 & \textbf{0.094 $\pm$ 0.003} \\
        \textbf{$\text{e}^2$IP (Ours)} & 0.471 $\pm$ 0.010  & 0.68 $\pm$ 0.02	&0.910 $\pm$ 0.024&\textbf{0.114 $\pm$ 0.028}& \textbf{1.31 $\pm$ 0.03} & 0.102 $\pm$ 0.016 \\
        \bottomrule
    \end{tabular}
\end{table*}
More importantly, probabilistic and calibration metrics reveal qualitative differences beyond point error.
DE(raw) is severely overconfident (reliability/PIT), leading to extremely low Coverage@95 and extremely large NLL, indicating that its raw uncertainty is not directly usable in practice.
DE(cal) reduces global overconfidence via isotropic scaling, but a single scalar cannot capture the configuration-dependent and orientation-specific error structure in condensed phases.
In contrast, $\text{e}^2$IP predicts a full covariance tensor end-to-end, explicitly modeling anisotropic and correlated uncertainty in 3D force space; this yields markedly better probabilistic consistency (lower NLL and CE/PIT) and fewer high-confidence-but-wrong predictions, making it more suitable for uncertainty-weighted simulation.
Fig.~\ref{fig:four} further shows stronger alignment between predicted epistemic scale and true force error, directly improving ranking-based decisions such as active learning~\cite{kulichenko2023uncertainty,jung2024active}.

To address whether $\text{e}^2$IP's advantage stems from modeling input-dependent aleatoric uncertainty, we additionally compare against MVE-DE, a heteroskedastic deep ensemble where each member outputs both force mean and diagonal variance. Despite having the capacity to model aleatoric uncertainty, MVE-DE achieves worse MAE (0.492) than $\text{e}^2$IP (0.471) and MAP-DE (0.386), and does not outperform $\text{e}^2$IP on any UQ metric (ES 0.73 vs 0.68; CE 0.139 vs 0.114; NLL 1.59 vs 1.31). The shallow ensemble (SE) performs even worse (NLL 15.2), suggesting that output-layer diversification alone is insufficient for reliable multivariate force UQ.

We note that NLL uses different distributional forms: Gaussian for DE/MVE-DE/SE, and Student-$t$ for $\text{e}^2$IP (the native predictive from NIW marginalization). The Energy Score (ES) is a strictly proper scoring rule that depends only on samples and requires no parametric assumptions~\cite{gneiting2007strictly}, making it directly comparable across distributional families. $\text{e}^2$IP achieves the best ES across all methods.

Overall, combining stronger probabilistic reliability (NLL, CE, ES), better error--uncertainty ranking alignment, and substantially lower inference cost (${\sim}4.5\times$ speedup over DE), $\text{e}^2$IP provides more actionable uncertainty while retaining competitive accuracy on \textit{ab initio} liquid water.

\subsection{Sample Efficiency with Equivariance}
\label{subsec:data_efficiency}

\begin{figure}[htbp]
    \centering
    \begin{minipage}{1.0\linewidth}
        \centering
        \includegraphics[width=1.0\linewidth]{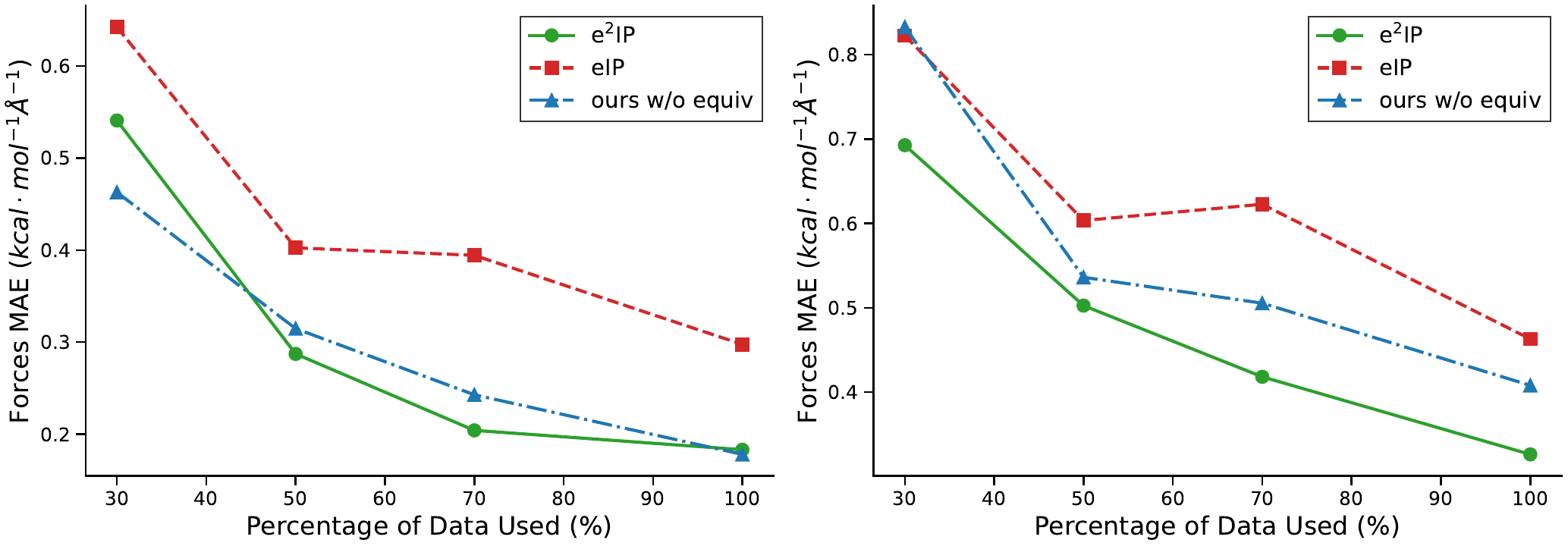}
    \end{minipage}

    \vspace{0.2cm}

    \begin{minipage}{1.0\linewidth}
        \centering
        \includegraphics[width=1.0\linewidth]{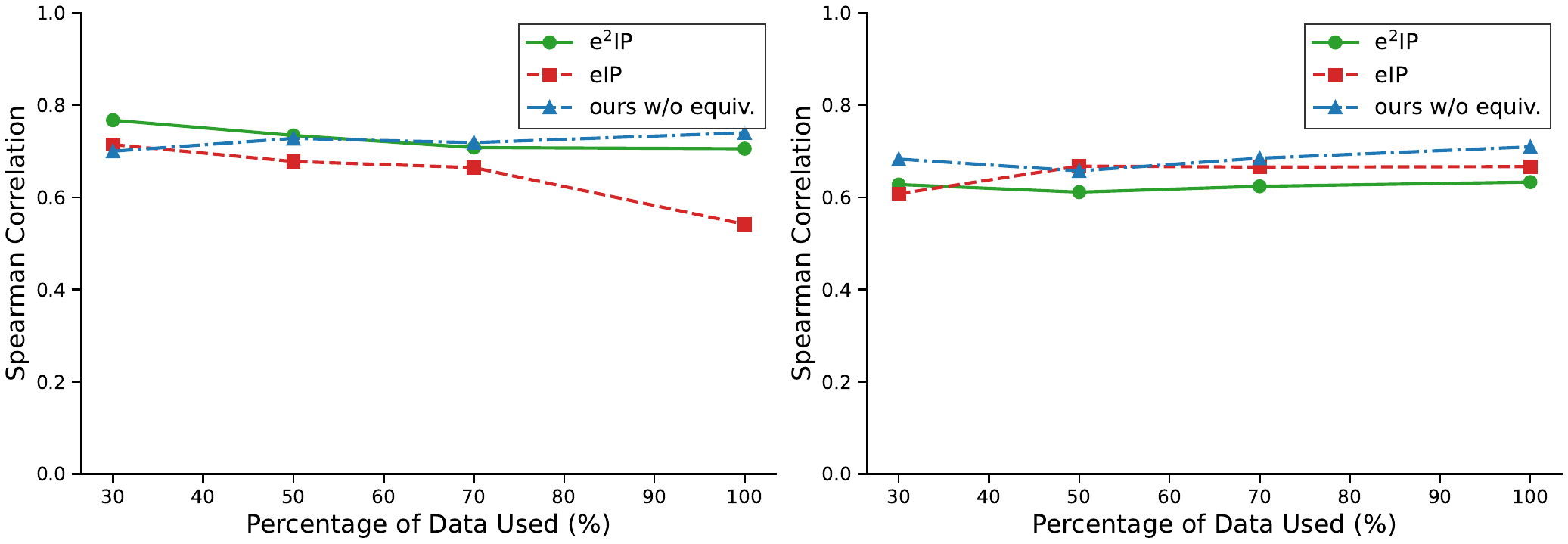}
    \end{minipage}

\caption{\textbf{Data efficiency on MD22 subsets.}
Force MAE (top; lower is better) and Spearman rank correlation (bottom; higher is better) on \textit{Buckyball-Catcher} (left) and \textit{DWCT} (right) as a function of training-set usage (30/50/70/100\%).
We compare $\text{e}^2$IP with eIP~\cite{xu2025evidential} and \textit{Ours w/o equiv.}.
$\text{e}^2$IP achieves consistently lower MAE across data regimes, with the largest gains on DWCT, while maintaining stable rank correlation.}
\label{fig:data_efficiency}

\end{figure}

We then evaluate sample efficiency on two molecular systems from the MD22 dataset, \textit{Buckyball-Catcher} ($C_{60}$--$C_{60}$) and \textit{Double-walled carbon nanotubes} (DWCT), which feature long-range and collective interactions and are more challenging than other MD22 molecules.
We compare $\text{e}^2$IP against two non-equivariant baselines: eIP~\cite{xu2025evidential} and \textit{Ours w/o equiv.}, an ablation that preserves the same likelihood objective (NLL) and uncertainty regularization but removes rotational equivariance by constructing a non-equivariant $\Sigma_0$ via a Cholesky parameterization.
All methods are trained using 30\%, 50\%, 70\%, and 100\% of the training data.
Fig.~\ref{fig:data_efficiency} reports force MAE and Spearman rank correlation as a function of training-set usage, capturing absolute accuracy and ranking quality, respectively.

Across both systems, $\text{e}^2$IP consistently exhibits improved data efficiency, with the largest and most uniform gains on DWCT.
This behavior aligns with the strong geometric anisotropy and extended interactions in nanotube systems, where enforcing rotational equivariance provides a powerful inductive bias that improves generalization under limited data.
Removing equivariance (\textit{Ours w/o equiv.}) degrades performance despite sharing the same training objective, indicating that the gains are not driven solely by the loss design.
On \textit{Buckyball-Catcher}, $\text{e}^2$IP also improves sample efficiency, although the margin is less pronounced.
We note that the non-equivariant variant can match or exceed $\text{e}^2$IP on raw accuracy at high data fractions on some systems; understanding when and why this occurs is an interesting open question.
Nevertheless, $\text{e}^2$IP maintains stable Spearman correlation as data decreases, suggesting that it preserves relative force ordering while improving accuracy. Importantly, the primary advantage of equivariance is not raw MAE dominance at every data point, but \emph{rotation-consistent uncertainty}: we further probe this by evaluating on randomly rotated DWCT test configurations (Table~\ref{tab:spearman_change_so3}), where $\text{e}^2$IP achieves substantially smaller $|\Delta\rho|$.

A physically meaningful force field should be \emph{objective} under rigid-body rotations: rotating the coordinates by $R$ should rotate the forces by the same $R$.
To explicitly probe this property, we evaluate robustness under \emph{proper} 3D rotations from the special orthogonal group
\[
SO(3)=\{R\in\mathbb{R}^{3\times 3}\mid R^\top R=I,\ \det(R)=+1\}.
\]
Given a configuration with coordinates of atom $i$ $\vec {x}_i$ and reference forces $\vec f_i$, we construct a rotated input $\vec{x}_i'=R\vec{x}_i$ and the correspondingly rotated labels $\vec{f}_i'=R\vec{f}_i$.
We then compute the force error magnitude
$e_i=\lVert \vec{f}_i'-\hat{\vec{f}}_i(\vec{x}_i')\rVert$, and an uncertainty score $u_i$ obtained from a rotationally invariant scalar proxy of the predicted uncertainty tensor (Eq.~\eqref{eq:scalar_proxy_main}).
To assess whether uncertainty remains \emph{actionable} after transformation, we measure the Spearman rank correlation between uncertainty and error, $\rho=\mathrm{Spearman}(\{u_i\},\{e_i\})$, report correlations on the original test set ($\rho_{\text{orig}}$) and the rotated test set ($\rho_{\text{rot}}$), and summarize the stability via $\Delta\rho=\rho_{\text{rot}}-\rho_{\text{orig}},$
where values closer to zero indicate better preservation of error--uncertainty alignment.

We focus on DWCT, where equivariance is expected to be particularly beneficial due to pronounced geometric anisotropy and extended interactions.
For each DWCT test configuration, we sample an independent random $R\in SO(3)$ to form a rotated test set, and compare $\text{e}^2$IP against eIP and the non-equivariant ablation \textit{Ours w/o equiv.}.
This setup isolates whether enforcing rotational equivariance improves transformation consistency and, critically, whether it preserves uncertainty-aware ranking under arbitrary 3D rotations.

Table~\ref{tab:spearman_change_so3} reports $\Delta\rho$ across different training-data regimes.
Applying random $SO(3)$ rotations causes a modest degradation in rank alignment for all methods, which is reasonable in practice: even when the target mapping is equivariant, finite numerical precision and common implementation choices (e.g., discretization, neighbor selection, and approximate geometric features) can introduce small rotation-dependent perturbations. We further verify that our method satisfies $SO(3)$ equivariance to
numerical precision; see Appendix~\ref{app:equivariance_validation} for details.
Importantly, the non-equivariant baseline eIP exhibits the largest negative shifts, indicating that its uncertainty ranking is most sensitive to rotations.
The ablation \textit{Ours w/o equiv.} is more stable than eIP, suggesting that matching the likelihood objective and uncertainty regularization helps, but it still shows noticeably larger changes than $\text{e}^2$IP.
Across all data regimes, $\text{e}^2$IP consistently achieves the smallest $|\Delta\rho|$ (Table~\ref{tab:spearman_change_so3}), demonstrating that enforcing rotational equivariance is crucial for maintaining uncertainty--error alignment under arbitrary 3D rotations and thus for reliable, uncertainty-driven ranking in downstream use.

\begin{table*}[htbp]
\centering
\caption{\textbf{Change in Spearman correlation under random $SO(3)$ rotations on DWCT.}
We report $|\Delta\rho|=|\rho_{\text{rot}}-\rho_{\text{orig}}|$ (closer to 0 is better). For multi-seed methods we report mean $\pm$ std. \textbf{Best results are highlighted in bold.}}
\label{tab:spearman_change_so3}
\setlength{\tabcolsep}{5pt}
\begin{tabular}{lccccc}
\toprule
\textbf{Method} & \textbf{30\%} & \textbf{50\%} & \textbf{70\%} & \textbf{100\%} & \textbf{Avg.} \\
\midrule
\textbf{DE (K=5)} & -- & -- & -- & 0.140 & -- \\
\textbf{eIP~\cite{xu2025evidential}} & 0.0856 & 0.1454 & 0.1302 & 0.077$\pm$0.044 & 0.1258 \\
\textbf{e$^2$IP (Ours) w/o equiv} & 0.0754 & 0.0384 & 0.0449 & 0.0432 & 0.0505 \\
\textbf{e$^2$IP (Ours)} & \textbf{0.0470} & \textbf{0.0235} & \textbf{0.0244} & \textbf{0.006$\pm$0.007} & \textbf{0.0239} \\
\bottomrule
\end{tabular}
\end{table*}

\subsection{Backbone-migration Evaluation on \textsc{Silica} Glass (OOD)}
\label{subsec:silica_ood}

To stress-test the backbone-migration capability of our uncertainty head, we run controlled experiments on the \textsc{Silica} glass benchmark~\cite{tan2023single} using another equivariant backbone \textsc{PaiNN}~\cite{schutt2021equivariant} as the base predictor.
We keep the \textsc{PaiNN} backbone and all training hyperparameters fixed across runs, and train different UQ heads with the \emph{same random seed} to minimize variance from initialization (full settings in Appendix~\ref{app:e2IP}).
A key aspect of this benchmark is that the official test split is intentionally \emph{out-of-distribution (OOD)}: training structures are generated under low temperature and low deformation rate, whereas test structures come from higher-temperature and higher-deformation-rate trajectories.
This shift probes whether a UQ method remains informative when generalization is genuinely difficult, rather than under an in-domain split.

Following the dataset protocol, we subsample 1000 structures from the available training pool for training and use the remaining training structures for validation; evaluation is performed on the OOD test set.
Table~\ref{tab:silica_backbone_migration} reports force MAE, rank correlation, and inference time. We additionally trained a DE baseline (K=5 PaiNN models) and re-ran eIP (5 seeds) and $\text{e}^2$IP (4 seeds) with error bars.
DE achieves the lowest MAE at $4.6\times$ inference cost (0.242 vs 0.053 ms/atom). $\text{e}^2$IP outperforms eIP on both MAE and Spearman, and achieves comparable Spearman to DE (0.775 vs 0.790) while maintaining single-model efficiency. This demonstrates that our uncertainty head transfers effectively to a different backbone and remains useful under a challenging distribution shift.

\begin{table}[H]
  \centering
  \caption{\textbf{Results on the \textsc{Silica} test set (OOD)}. Force MAE, Spearman $\rho$, and inference time. Mean $\pm$ std over seeds. \textbf{Best} in bold.}
  \label{tab:silica_backbone_migration}
  \footnotesize
  \setlength{\tabcolsep}{2pt}
  \begin{tabular}{lccc}
    \toprule
    \textbf{Method} & \textbf{MAE}$\downarrow$ & \textbf{$\rho$}$\uparrow$ & \textbf{ms/atom}$\downarrow$ \\
    \midrule
    eIP (5s)~\cite{xu2025evidential} & 1.622{\scriptsize$\pm$.040} & .767{\scriptsize$\pm$.002} & \textbf{.049{\scriptsize$\pm$.003}} \\
    DE (K=5) & \textbf{1.059} & \textbf{.790} & .242{\scriptsize$\pm$.010} \\
    e$^2$IP (4s) & 1.559{\scriptsize$\pm$.065} & .775{\scriptsize$\pm$.008} & .053{\scriptsize$\pm$.002} \\
    \bottomrule
  \end{tabular}
\end{table}

\subsection{Evaluation on rMD17 Molecular Force Fields}
\label{subsec:rmd17}

To evaluate on standard molecular force-field benchmarks with diverse molecular sizes, we compare DE, eIP, $\text{e}^2$IP, and MVE-DE on two molecules from the revised MD17 dataset~\cite{christensen2020role}: aspirin (21 atoms) and azobenzene (24 atoms). All methods use the same AlphaNet backbone and hyperparameters. Results are reported over multiple seeds (Table~\ref{tab:rmd17_results}).

\begin{table}[H]
  \centering
  \caption{\textbf{Results on rMD17.} MAE, ES, CE, NLL. Mean $\pm$ std. \textbf{Best} in bold.}
  \label{tab:rmd17_results}
  \setlength{\tabcolsep}{2pt}
  \footnotesize
  \begin{tabular}{llcccc}
    \toprule
    & \textbf{Method} & \textbf{MAE}$\downarrow$ & \textbf{ES}$\downarrow$ & \textbf{CE}$\downarrow$ & \textbf{NLL}$\downarrow$ \\
    \midrule
    \multirow{4}{*}{\rotatebox{90}{Asp.}}
      & DE(cal) & \textbf{.187} & .703 & .423 & 3.30 \\
      & MVE-DE & .496 & .830 & .444 & 3.35 \\
      & eIP & .299{\scriptsize$\pm$.024} & .453{\scriptsize$\pm$.033} & .110{\scriptsize$\pm$.012} & 2.43{\scriptsize$\pm$.07} \\
      & e$^2$IP & .272{\scriptsize$\pm$.008} & \textbf{.395{\scriptsize$\pm$.010}} & \textbf{.079{\scriptsize$\pm$.037}} & \textbf{.83{\scriptsize$\pm$.05}} \\
    \midrule
    \multirow{4}{*}{\rotatebox{90}{Azo.}}
      & DE(cal) & .139 & .341 & .327 & 2.34 \\
      & MVE-DE & .640 & 1.238 & .472 & 4.93 \\
      & eIP & .217{\scriptsize$\pm$.115} & .323{\scriptsize$\pm$.165} & .149{\scriptsize$\pm$.050} & .35{\scriptsize$\pm$1.28} \\
      & e$^2$IP & \textbf{.128{\scriptsize$\pm$.020}} & \textbf{.187{\scriptsize$\pm$.028}} & \textbf{.076{\scriptsize$\pm$.052}} & \textbf{$-$1.47{\scriptsize$\pm$.15}} \\
    \bottomrule
  \end{tabular}
\end{table}

On both molecules, $\text{e}^2$IP achieves the best CE, ES, and NLL. On azobenzene, $\text{e}^2$IP also achieves the lowest MAE (0.128 vs DE's 0.139), demonstrating that the evidential framework can compete with ensembles on point accuracy.
MVE-DE performs substantially worse than both MAP-DE and $\text{e}^2$IP across all metrics, with significantly degraded MAE (0.496 on aspirin, 0.640 on azobenzene) and poor calibration (CE $>$ 0.44). Furthermore, some MVE-DE seeds failed to converge during training (aspirin seed 1, azobenzene seeds 3--4), indicating training instability when applying heteroskedastic Gaussian NLL to autograd-derived forces --- a known challenge in MLIP UQ.
These results, combined with the liquid water experiments, consistently demonstrate that $\text{e}^2$IP's NIW evidential framework provides a more effective joint force--uncertainty parameterization than heteroskedastic ensembles.

\subsection{Ablation Study: Necessity of Spectral Stabilization}
\label{subsec:ablation_damping}
\begin{figure}[htbp]
    \centering
    
    \includegraphics[width=1.0\linewidth]{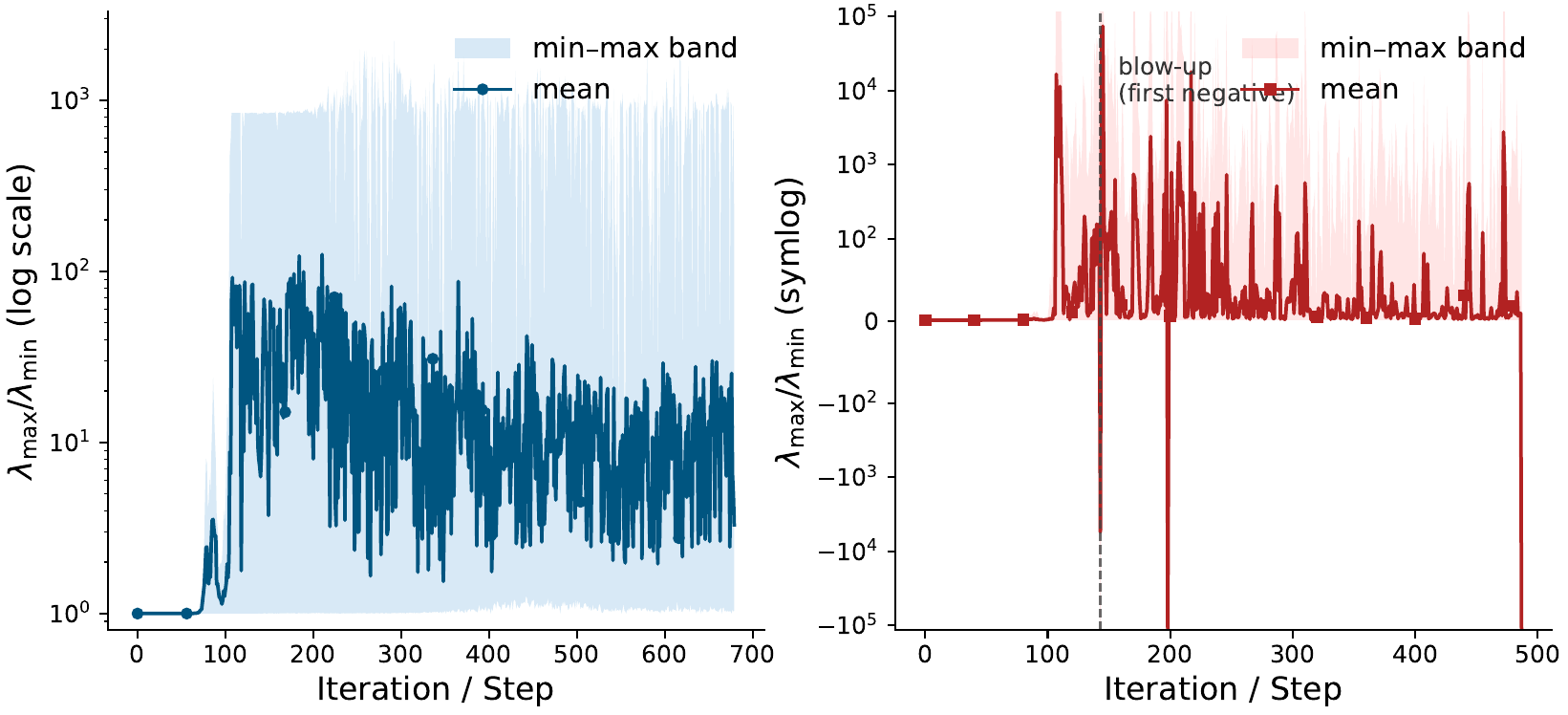}
    \caption{\textbf{Spectral stabilization is necessary for numerical robustness.}
We track the batch-wise mean and min--max band of the eigenvalue ratio $\lambda_{\max}(\Sigma_0)/\lambda_{\min}(\Sigma_0)$ during early training.
\emph{left (w/ damper):} the ratio stays bounded and training remains stable.
\emph{right (w/o damper):} the ratio rapidly explodes and becomes erratic; negative values indicate that $\lambda_{\min}(\Sigma_0)$ turns non-positive due to finite-precision effects, after which Cholesky factorization fails (dashed line).}
\label{fig:condnum_ablation}
\end{figure}
Our method constructs the base covariance $\Sigma_0$ during training, which must remain positive definite and numerically well-conditioned to support stable optimization (e.g., Cholesky factorization).
To isolate the role of the proposed damper, we ablate the spectral stabilization while keeping all other components unchanged on the liquid water dataset from Sec.~\ref{subsec:water_experiment}.
The collapse behavior without spectral stabilization is consistently observed across other benchmarks, suggesting that the instability is a general numerical issue.

Figure~\ref{fig:condnum_ablation} monitors the batch-wise mean and min--max band of the condition proxy $\lambda_{\max}(\Sigma_0)/\lambda_{\min}(\Sigma_0)$ over early training.
With damping (left), the ratio may increase during initial adaptation but quickly enters a bounded regime, and the min--max band remains controlled, indicating that the covariance spectrum stays well-conditioned throughout training.

Without damping (right), the spectrum becomes unstable: the eigenvalue ratio grows by orders of magnitude and fluctuates sharply across samples, signaling severe ill-conditioning.
Moreover, the ratio can become negative, which implies $\lambda_{\min}(\Sigma_0)\le 0$ and therefore a loss of positive definiteness under finite precision.
After the first such event (dashed line), Cholesky factorization fails and training diverges.
Overall, this ablation shows that spectral stabilization is not merely a convenience but a necessary component for reliable covariance construction in practice.

\section{Discussion and Conclusion}

\paragraph{Discussion.}
$\text{e}^{2}$IP couples evidential learning with $SE(3)$-equivariant rank-2 tensor modeling: a tangent-space parameterization maps unconstrained symmetric matrices to SPD covariances via the Riemannian exponential, guaranteeing positive definiteness and rotation consistency by construction.
The result is a full $3\times3$ uncertainty tensor that captures directional structure inaccessible to diagonal formulations.
Our ablations show that spectral damping is essential for stable full-covariance learning (Fig.~\ref{fig:condnum_ablation}), and our comparison with heteroskedastic deep ensembles (MVE-DE) demonstrates that the NIW framework achieves superior calibration without the training instability observed in Gaussian NLL-based approaches.
Because the uncertainty head operates on equivariant features and outputs structured tensor components, it integrates with different equivariant backbones (AlphaNet, PaiNN) with minimal modification.

\paragraph{Limitations and Future Work.}
Our evaluation primarily targets $SO(3)$ consistency; extending to $O(3)$ (including reflections) and stress-testing under more extreme non-equilibrium conditions are natural next steps.
In addition, the Normal-Inverse-Wishart prior may be restrictive under heavy-tailed residuals; richer priors (e.g., mixtures or robust alternatives) could further improve reliability.
An important direction is to quantify downstream gains in uncertainty-aware molecular simulation, such as end-to-end active learning loops and stability improvements in long-timescale MD simulations.

\paragraph{Conclusion.}
We presented $\text{e}^{2}$IP, an $SE(3)$-equivariant evidential framework that models aleatoric and epistemic uncertainty in atomic forces via full SPD covariances.
Across molecular dynamics benchmarks, $\text{e}^{2}$IP consistently outperforms baselines in uncertainty reliability and calibration quality, while achieving competitive predictive accuracy at single-model inference cost.

These results suggest that incorporating symmetry and tensor structure into uncertainty modeling is a principled and practical alternative to ensembling for uncertainty-aware molecular simulation.

\bibliography{example_paper}
\bibliographystyle{icml2026}

\section{Impact Statements}
This paper presents work whose goal is to advance the field of Machine Learning. We do not anticipate immediate negative societal impacts.

\newpage
\appendix
\onecolumn
\section{Additional details for $\text{e}^2$IP}
\label{app:e2IP}
Table \ref{tab:hyperparams} summarizes the hyperparameters used for the AlphaNet and PAINN backbones in our experiments. Both models share the same optimization strategy and loss weighting scheme.

\begin{table}[h]
    \centering
    \caption{Hyperparameters for AlphaNet and PAINN models used in experiments.}
    \label{tab:hyperparams}
    \begin{tabular}{llcc}
        \toprule
        \textbf{Category} & \textbf{Hyperparameter} & \textbf{AlphaNet} & \textbf{PAINN} \\
        \midrule
        \multirow{6}{*}{\textbf{Optimization}} 
          & Optimizer & \multicolumn{2}{c}{AdamW} \\
          & Learning Rate & \multicolumn{2}{c}{$2 \times 10^{-4}$} \\
          & Betas & \multicolumn{2}{c}{$(0.9, 0.999)$} \\
          & Weight Decay & \multicolumn{2}{c}{$0.0$} \\
          & Scheduler & \multicolumn{2}{c}{ReduceLROnPlateau (min)} \\
          & Factor / Patience & \multicolumn{2}{c}{$0.85$ / $50$} \\
        \midrule
        \multirow{7}{*}{\textbf{Architecture}} 
          & Number of Layers & $4$ & $6$ \\
          & Cutoff Radius ($r_c$) & $5.0$ \AA & $5.0$ \AA \\
          & Hidden Channels & $128$ & $128$ \\
          & Num. Radial Basis & $64$ & 128 \\
          & Main $\chi_1$ / MP $\chi_1$ & $32$ / $32$ & -- \\
          & $\chi_2$ & $8$ & -- \\
          & Hidden Channels ($\chi$) & $96$ & -- \\
        \midrule
        \multirow{3}{*}{\textbf{Loss \& Reg.}} 
          & Energy Weight & \multicolumn{2}{c}{$1.0$} \\
          & Force Weight & \multicolumn{2}{c}{$10000.0$} \\
          & Evidence Reg. ($\lambda_{\text{reg}}$) & \multicolumn{2}{c}{$0.1$} \\
        \bottomrule
    \end{tabular}
\end{table}
\section{Notation and validity conditions}
\label{app:notation}

Table~\ref{tab:notation_detailed} summarizes symbols used in $\text{e}^2$IP.

\begin{longtable}{l l p{10cm}} 
    
    \caption{Detailed Notation and Physical Interpretations used in $e^2$IP.} \label{tab:notation_detailed} \\
    \toprule
    \textbf{Symbol} & \textbf{Mathematical Name} & \textbf{Physical Interpretation \& Intuition} \\
    \midrule
    \endfirsthead
    
    \multicolumn{3}{c}{{\bfseries \tablename\ \thetable{} -- continued from previous page}} \\
    \toprule
    \textbf{Symbol} & \textbf{Mathematical Name} & \textbf{Physical Interpretation \& Intuition} \\
    \midrule
    \endhead
    
    \midrule
    \multicolumn{3}{r}{{Continued on next page...}} \\
    \endfoot
    
    \bottomrule
    \endlastfoot
    
    
    \multicolumn{3}{l}{\textit{\textbf{Input, Target \& Latent Variables}}} \\
    $\vec{y}_i \in \mathbb{R}^3$ & Target Vector & The ground truth atomic force derived from DFT calculations on atom $i$. \\
    $\vec{\mu}_i \in \mathbb{R}^3$ & Latent Mean & The unknown "true" mean force. In the evidential framework, this is treated as a random variable following a Normal distribution. \\
    $\Sigma_i \in \mathbb{S}_{++}^3$ & Latent Covariance & The unknown "true" covariance of the aleatoric noise. It is treated as a random variable following an Inverse-Wishart distribution. \\
    $d$ & Dimension & The dimensionality of the target vector ($d=3$ for atomic forces). \\
    
    \midrule
    \multicolumn{3}{l}{\textit{\textbf{Evidential Parameters (Network Outputs)}}} \\
    $\vec{\gamma}_i \in \mathbb{R}^3$ & Evidential Mean & The model's best prediction for the atomic force. It serves as the mean of the predictive Student-t distribution. \\
    $\Sigma_{0,i} \in \mathbb{S}_{++}^3$ & Nominal Covariance & A symmetric positive definite matrix determining the shape of the aleatoric uncertainty. It captures directional correlations in force noise. \\
    $\nu_i \in \mathbb{R}^+$ & Degrees of Freedom of IW distribution  & Represents the quantity of evidence supporting the estimated covariance shape $\Sigma_{0,i}$. Higher $\nu_i$ means the model is more confident about the aleatoric noise profile. \\
    $\kappa_i \in \mathbb{R}^+$ & Belief Count & Represents the quantity of evidence supporting the mean prediction $\vec{\gamma}_i$. Higher $\kappa_i$ implies lower epistemic uncertainty (i.e., the model has seen ample similar training data). \\
    $\Psi_i \in \mathbb{S}_{++}^3$ & Scale Matrix & The scale matrix for the Inverse-Wishart prior, reparameterized as $\Psi_i = \nu_i \Sigma_{0,i}$ for physical interpretability. \\
    
    \midrule
    \multicolumn{3}{l}{\textit{\textbf{Uncertainty Quantification Measures}}} \\
    $U_{ale} \in \mathbb{S}_{++}^3$ & Aleatoric Uncertainty & Expected data noise ($\mathbb{E}[\Sigma_i]$). It captures irreducible variability (e.g., thermal fluctuations) and transforms as a rank-2 tensor. \\
    $U_{epi} \in \mathbb{S}_{++}^3$ & Epistemic Uncertainty & Model uncertainty ($\text{Var}[\vec{\mu}_i]$). It captures the lack of knowledge in OOD regions. Computed as $\frac{\nu_i \Sigma_{0,i}}{\kappa_i(\nu_i - d - 1)}$. \\
    $m_i$ & DoF of t-distribution & Degrees of freedom for the posterior Student-t distribution, defined as $m_i = \nu_i - d + 1$. \\
    
    \midrule
    \multicolumn{3}{l}{\textit{\textbf{Geometry, Tangent Space \& Optimization}}} \\
    $\mathcal{S}_i \in \mathrm{sym}(3)$ & Tangent Vector & A symmetric matrix in the tangent space of $\mathbb{S}_{++}^3$ at the identity. The network predicts this unbounded tensor to ensure unconstrained optimization. \\
    $\exp(\mathcal{S}_i)$ & Riemannian Exponential Map & The map $\mathrm{sym}(3) \to \mathbb{S}_{++}^3$ that transforms $\mathcal{S}_i$ into the valid SPD covariance $\Sigma_{0,i}$, guaranteeing positive definiteness and equivariance by construction~\cite{moakher2005differential,arsigny2007geometric}. \\
    $\alpha_i$ & Spectral Damper & A scalar coefficient used to damp the magnitude of the traceless part of $\mathcal{S}_i$ during training, preventing numerical instability (ill-conditioning). \\
    $\mathcal{L}_{NLL}$ & Negative Log-Likelihood & The primary loss function based on the multivariate Student-t distribution, encouraging accuracy and calibration. \\
    $\mathcal{L}_{reg}$ & Evidence Regularizer & A penalty term that discourages the model from assigning high evidence ($\nu_i, \kappa_i$) to data points with large prediction errors. \\
    
\end{longtable}

\paragraph{Validity conditions.}
The predictive Student-$t$ has positive degrees of freedom when $m_i=\nu_i-d+1>0$, i.e., $\nu_i>d-1$.
In the main text we additionally require $\nu_i>d+1$ so that $\mathbb{E}[\mathbf{\Sigma}_i]$ exists and the tensor
uncertainty decomposition in Eq.~\eqref{eq:ale_epi_main} is well-defined.

\section{Derivation of the multivariate Student-\textit{t} NLL}
\label{app:student_t_nll}

This appendix derives the expanded negative log-likelihood used in Eq.~\eqref{eq:loss} in the main text,
starting from the NIW predictive distribution.

\subsection{Predictive Student-\textit{t} from an NIW prior}
\label{app:student_t_predictive}

Let $\vec{y}_i \in \mathbb{R}^d$ denote the target vector and let the likelihood be
\begin{equation}
p(\vec{y}_i \mid \vec{\mu}_i, \mathbf{\Sigma}_i)
= \mathcal{N}(\vec{y}_i \mid \vec{\mu}_i, \mathbf{\Sigma}_i).
\end{equation}
We place a Normal--Inverse--Wishart (NIW) prior on $(\vec{\mu}_i,\mathbf{\Sigma}_i)$ with parameters
$(\vec{\gamma}_i, \kappa_i, \mathbf{\Psi}_i, \nu_i)$:
\begin{equation}
\mathbf{\Sigma}_i \sim \mathcal{W}^{-1}(\mathbf{\Psi}_i,\nu_i),
\qquad
\vec{\mu}_i \mid \mathbf{\Sigma}_i \sim \mathcal{N}\!\left(\vec{\gamma}_i, \frac{1}{\kappa_i}\mathbf{\Sigma}_i\right),
\end{equation}
where $\kappa_i>0$ and $\nu_i>d-1$ so that the predictive degrees of freedom are positive.

Marginalizing $(\vec{\mu}_i,\mathbf{\Sigma}_i)$ yields a multivariate Student-\textit{t} predictive distribution:
\begin{equation}
\label{eq:app_student_t}
p(\vec{y}_i \mid \vec{\gamma}_i, \kappa_i, \mathbf{\Psi}_i, \nu_i)
=
St_{\nu_i-d+1}\!\left(
\vec{y}_i \,\middle|\,
\vec{\gamma}_i,\,
\frac{1+\kappa_i}{\kappa_i(\nu_i-d+1)}\,\mathbf{\Psi}_i
\right).
\end{equation}

\subsection{NLL in a log-determinant form}
\label{app:nll_logdet}

An equivalent closed form for the negative log-likelihood is
\begin{equation}
\begin{split}
\mathcal{L}_{\mathrm{NLL},i}
&= \log\Gamma\!\left(\frac{\nu_i-d+1}{2}\right)
 - \log\Gamma\!\left(\frac{\nu_i+1}{2}\right)
 + \frac{d}{2}\log\!\left(\pi\frac{1+\kappa_i}{\kappa_i}\right)
 - \frac{\nu_i}{2}\log\left|\mathbf{\Psi}_i\right| \\
&\quad
 + \frac{\nu_i+1}{2}\log\left|
\mathbf{\Psi}_i + \frac{\kappa_i}{1+\kappa_i}
(\vec{y}_i-\vec{\gamma}_i)(\vec{y}_i-\vec{\gamma}_i)^\top
\right|.
\end{split}
\label{eq:app_nll_psy}
\end{equation}

\subsection{Reparameterization $\mathbf{\Psi}_i=\nu_i\mathbf{\Sigma}_{0,i}$}
\label{app:psi_reparam}

Following the main text (Eq.~\eqref{eq:psi_reparam_main}), define $\mathbf{\Psi}_i \triangleq \nu_i \mathbf{\Sigma}_{0,i}$ with
$\mathbf{\Sigma}_{0,i}$ being a SPD matrix. Substituting into Eq.~\eqref{eq:app_nll_psy} gives
\begin{equation}
\begin{split}
\mathcal{L}_{\mathrm{NLL},i}
&= \log\Gamma\!\left(\frac{\nu_i-d+1}{2}\right)
 - \log\Gamma\!\left(\frac{\nu_i+1}{2}\right)
 + \frac{d}{2}\log\!\left(\nu_i\pi\frac{1+\kappa_i}{\kappa_i}\right)
 - \frac{\nu_i}{2}\log\left|\mathbf{\Sigma}_{0,i}\right| \\
&\quad
 + \frac{\nu_i+1}{2}\log\left|
\mathbf{\Sigma}_{0,i} + \frac{\kappa_i}{\nu_i(1+\kappa_i)}
(\vec{y}_i-\vec{\gamma}_i)(\vec{y}_i-\vec{\gamma}_i)^\top
\right|.
\end{split}
\label{eq:app_nll_sigma0_pre}
\end{equation}

\subsection{Determinant lemma and simplification}
\label{app:det_lemma_simplify}

Define $\vec{v}_i \triangleq \vec{y}_i-\vec{\gamma}_i$ and
$\alpha_i \triangleq \frac{\kappa_i}{\nu_i(1+\kappa_i)}$.
We apply the matrix determinant lemma:
\begin{equation}
\left|\mathbf{A} + \mathbf{u}\mathbf{u}^\top\right|
= |\mathbf{A}|\left(1+\mathbf{u}^\top \mathbf{A}^{-1}\mathbf{u}\right),
\qquad \text{for invertible } \mathbf{A}.
\label{eq:det_lemma}
\end{equation}
Setting $\mathbf{A}=\mathbf{\Sigma}_{0,i}$ and $\mathbf{u}=\sqrt{\alpha_i}\,\vec{v}_i$ yields
\begin{equation}
\left|\mathbf{\Sigma}_{0,i} + \alpha_i \vec{v}_i\vec{v}_i^\top\right|
=
\left|\mathbf{\Sigma}_{0,i}\right|
\left(1+\alpha_i\,\vec{v}_i^\top \mathbf{\Sigma}_{0,i}^{-1}\vec{v}_i\right).
\end{equation}
Taking logarithms,
\begin{equation}
\log\left|\mathbf{\Sigma}_{0,i} + \alpha_i \vec{v}_i\vec{v}_i^\top\right|
=
\log\left|\mathbf{\Sigma}_{0,i}\right|
+
\log\!\left(1+\alpha_i\,\vec{v}_i^\top \mathbf{\Sigma}_{0,i}^{-1}\vec{v}_i\right).
\label{eq:logdet_split}
\end{equation}
Substituting Eq.~\eqref{eq:logdet_split} into Eq.~\eqref{eq:app_nll_sigma0_pre}, the
$\log|\mathbf{\Sigma}_{0,i}|$ terms combine as
\begin{equation}
-\frac{\nu_i}{2}\log|\mathbf{\Sigma}_{0,i}|
+\frac{\nu_i+1}{2}\log|\mathbf{\Sigma}_{0,i}|
=
\frac{1}{2}\log|\mathbf{\Sigma}_{0,i}|.
\end{equation}
Therefore we obtain the expanded form used in the main text:
\begin{equation}
\begin{split}
\mathcal{L}_{\mathrm{NLL},i}
&= \log\Gamma\!\left(\frac{\nu_i-d+1}{2}\right)
 - \log\Gamma\!\left(\frac{\nu_i+1}{2}\right)
 + \frac{d}{2}\log\!\left(\frac{\pi\nu_i(1+\kappa_i)}{\kappa_i}\right)
 + \frac{1}{2}\log\left|\mathbf{\Sigma}_{0,i}\right| \\
&\quad
 + \frac{\nu_i+1}{2}\log\!\left(
1 + \frac{\kappa_i}{\nu_i(1+\kappa_i)}
(\vec{y}_i-\vec{\gamma}_i)^\top \mathbf{\Sigma}_{0,i}^{-1}(\vec{y}_i-\vec{\gamma}_i)
\right).
\end{split}
\label{eq:app_nll_final}
\end{equation}

\section{Irrep-to-Cartesian map and the constant basis matrix $\mathbf{Q}$}
\label{app:Q_map}

This appendix specifies the implementation-dependent details omitted in
Sec.~\ref{subsec:parameterization}. Let $\mathbf{z}_i=[s_i,\mathbf{t}_i]\in\mathbb{R}^{1+5}$ denote the coefficients
in $(0e\oplus2e)$. We map $\mathbf{z}_i$ to a Cartesian symmetric matrix $\mathcal{S}_i\in\mathrm{sym}(3)$ using a
fixed Clebsch--Gordan basis.

\paragraph{One possible convention.}
We first produce a (not-necessarily symmetric) $3\times 3$ matrix $\mathbf{S}_i$ by a linear map,
\begin{equation}
\mathrm{vec}(\mathbf{S}_i) = \mathbf{Q}^\top \mathbf{z}_i,
\qquad
\mathbf{Q}\in\mathbb{R}^{9\times 6},
\end{equation}
where $\mathrm{vec}(\cdot)$ stacks entries in a fixed order (e.g., row-major).
We then symmetrize:
\begin{equation}
\mathcal{S}_i \triangleq \tfrac{1}{2}(\mathbf{S}_i+\mathbf{S}_i^\top)\in\mathrm{sym}(3).
\end{equation}
(Any equivalent convention that yields $\mathcal{S}_i$ transforming as a rank-2 tensor is valid; we report the exact
convention used in code for reproducibility.)

\paragraph{Equivariance.}
By construction of $\mathbf{Q}$ from CG coefficients of $(1o\otimes 1o)\to (0e\oplus2e)$, the resulting
$\mathcal{S}_i$ transforms as $\mathcal{S}_i\mapsto R\mathcal{S}_iR^\top$ for $R\in SO(3)$.
\paragraph{Implementation details.}
We utilized the \texttt{e3nn.io.CartesianTensor} method for this construction. Due to performance inefficiencies in the original source code, we re-implemented the transformation method and validated the consistency of our results against the original implementation.

\section{Damping functions used for spectral stabilization}
\label{app:damping_details}

For completeness we provide the explicit damping used in Section~\ref{subsec:damping}.

\paragraph{Linear--Tanh damper.}
We define a smooth piecewise function that is linear near the origin and saturates to a ceiling:
\begin{equation}
\phi_{\mathrm{high}}(x) =
\begin{cases}
x, & |x| \le \tau, \\
\mathrm{sgn}(x)\left[(c - \tau)\tanh\!\left(\frac{|x| - \tau}{c - \tau}\right) + \tau\right], & |x| > \tau,
\end{cases}
\end{equation}
with threshold $\tau$ and ceiling $c$. The lower damper $\phi_{\mathrm{low}}$ is defined analogously to enforce a
floor on the isotropic channel. In our experiments, we use $\tau=4$ and $c=5$.
\begin{figure}[t]
\centering
\begin{tikzpicture}
\begin{axis}[
    width=0.75\linewidth,
    height=0.5\linewidth,
    axis lines=middle,
    xlabel={$x$},
    ylabel={$\phi_{\mathrm{high}}(x)$},
    xmin=-7, xmax=7,
    ymin=-6, ymax=6,
    samples=300,
    domain=-7:7,
    legend style={at={(0.02,0.98)},anchor=north west},
]

\def\tau{4.0}
\def\cval{5.0}

\addplot[thick, blue]
    {abs(x)<=\tau ? x :
     sign(x)*((\cval-\tau)*tanh((abs(x)-\tau)/(\cval-\tau))+\tau)};

\addlegendentry{Linear--Tanh damper}

\addplot[dashed, gray] coordinates {(\tau,-4) (\tau,4)};
\addplot[dashed, gray] coordinates {(-\tau,-4) (-\tau,4)};

\addplot[dashed, gray] coordinates {(-5,\cval) (5,\cval)};
\addplot[dashed, gray] coordinates {(-5,-\cval) (5,-\cval)};

\end{axis}
\end{tikzpicture}
\caption{Linear--Tanh damping function with threshold $\tau=4$ and ceiling $c=5$.
The mapping is linear near the origin and smoothly saturates for large inputs.}
\label{fig:linear_tanh_damper}
\end{figure}
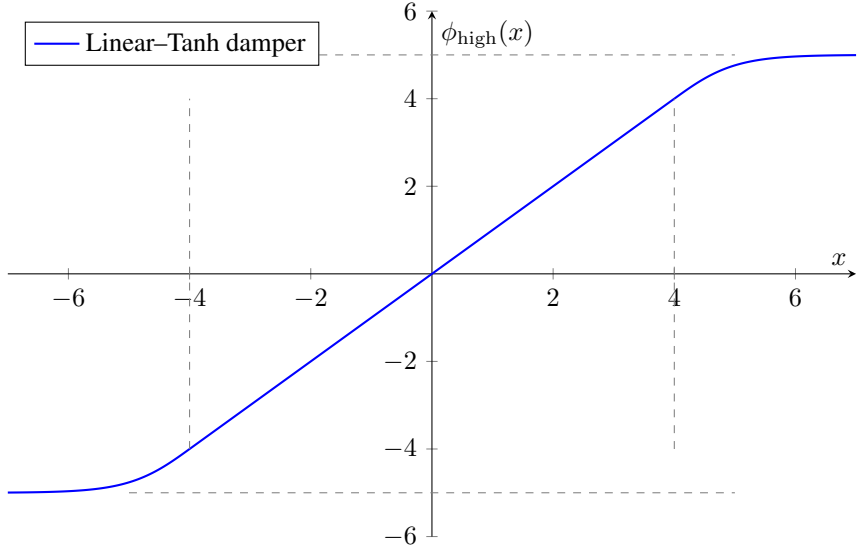

\paragraph{Coefficient-space application.}
Scalar: $\tilde{s}_i=\phi_{\mathrm{low}}(\phi_{\mathrm{high}}(s_i))$.
Vector: $\tilde{\mathbf{t}}_i=\alpha_i\mathbf{t}_i$ with $\alpha_i$ defined in Eq.~\eqref{eq:l2_damp_main}.
\section{Evaluation Metrics}
\label{app:metrics}

We evaluate uncertainty quality for vector-valued force prediction using distribution-aware calibration diagnostics together with negative log-likelihood (NLL) and the multivariate Energy Score (ES). Each atom-level force is a 3D vector. For sample $i$, let the target and prediction be $y_i\in\mathbb{R}^3$ and $\mu_i\in\mathbb{R}^3$, and define the error
\begin{equation}
e_i = y_i - \mu_i \in \mathbb{R}^3.
\end{equation}
Let $\Sigma_i\in\mathbb{R}^{3\times 3}$ denote the predicted positive definite matrix used by the model.
For the Gaussian baseline, $\Sigma_i$ is the predictive covariance. For our Student-$t$ model, we use the
scale matrix $\Lambda_i$ defined as
\begin{equation}
\Lambda_i = \frac{(1+\kappa)\nu_i}{\kappa(\nu_i-d+1)}\Sigma_i,
\qquad m_i := \nu_i-d+1.
\end{equation}

\paragraph{Predictive families.}
Our method (e$^2$IP) outputs a multivariate Student-$t$ predictive distribution with effective degrees of freedom $m_i$ and scale matrix $\Lambda_i$:
\begin{equation}
e_i \sim St_{m_i}\!\left(0,\Lambda_i\right).
\end{equation}
The deep ensemble baseline uses a Gaussian predictive distribution
\begin{equation}
e_i \sim \mathcal{N}(0,\Sigma_i).
\end{equation}

where $\Sigma_i$ is constructed from epistemic covariance plus an isotropic floor as described in the main text.

\subsection{Distribution-aware Calibration Curve, Coverage, and Calibration Error}
\label{app:calibration}

Calibration is assessed via the probability integral transform (PIT). For each method, we map the multivariate residual $e_i$ to a scalar $u_i\in(0,1)$ such that $u_i$ is uniform when the predictive distribution is calibrated.

\paragraph{Gaussian baseline.}
For the ensemble Gaussian predictive model, define the squared Mahalanobis distance
\begin{equation}
m_i^2 = e_i^\top \Sigma_i^{-1} e_i.
\end{equation}
If $e_i\sim \mathcal{N}(0,\Sigma_i)$ in 3D, then $m_i^2\sim \chi^2_3$, and we set
\begin{equation}
u_i^{\text{Gauss}} = F_{\chi^2_3}(m_i^2),
\end{equation}

where $F_{\chi^2_3}$ denotes the CDF of $\chi^2_3$. Under perfect calibration, $u_i^{\text{Gauss}}\sim\mathrm{Unif}(0,1)$.

\paragraph{Student-$t$ model (ours).}
For our Student-$t$ predictive model in 3D, we again use the quadratic form $m_i^2=e_i^\top \Lambda_i^{-1}e_i$. If $e_i\sim t_{\nu_i-d+1}(0,\frac{(1+\kappa)\nu}{\kappa(\nu-d+1)}\Sigma_i)$, then the scaled statistic
\begin{equation}
x_i = \frac{m_i^2}{3}
\end{equation}
follows an $F$ distribution:
\begin{equation}
x_i \sim F(3,\nu_i-d+1).
\end{equation}
We therefore define the PIT for our method as
\begin{equation}
u_i^{t} = F_{F(3,\nu_i-d+1)}\!\left(\frac{m_i^2}{3}\right),
\end{equation}
where $F_{F(3,\nu_i-d+1)}(\cdot)$ is the CDF of the $F(3,\nu_i)$ distribution. Under perfect calibration, $u_i^{t}\sim\mathrm{Unif}(0,1)$.

\paragraph{Unified calibration curve.}
Given a grid of nominal confidence levels $\mathcal{P}=\{p_1,\dots,p_M\}\subset(0,1)$, we define the empirical (observed) coverage curve for any method as
\begin{equation}
\mathrm{Obs}(p) = \frac{1}{N}\sum_{i=1}^{N}\mathbb{I}[u_i \le p],\qquad p\in\mathcal{P},
\end{equation}
where $u_i$ is the PIT scalar computed using the method's own predictive family (Gaussian or Student-$t$). The ideal calibration curve satisfies $\mathrm{Obs}(p)=p$.

\paragraph{Coverage at a nominal level.}
We report $\mathrm{Coverage@}p^\star=\mathrm{Obs}(p^\star)$ for selected nominal levels (e.g., $p^\star\in\{0.8,0.9,0.95\}$). Values closer to $p^\star$ indicate better calibration at that confidence level.

\paragraph{Calibration error.}
We summarize deviation from the ideal curve using an $\ell_1$ calibration error over the grid:
\begin{equation}
\mathrm{CE}_{\ell_1} = \frac{1}{M}\sum_{m=1}^{M}\left|\mathrm{Obs}(p_m)-p_m\right|.
\end{equation}

\subsection{Negative Log-Likelihood (Model-specific)}
\label{app:nll}

We report the average negative log-likelihood under each method's predictive distribution:
\begin{equation}
\mathrm{NLL} = -\frac{1}{N}\sum_{i=1}^{N}\log p(e_i),
\end{equation}
where $p(e_i)$ is the density of $\mathcal{N}(0,\Sigma_i)$ for the ensemble baseline and the density of $t_{m_i}(0,\Lambda_i)$ for our method.
Since these correspond to different distribution families, NLL should be interpreted as a model-specific proper score.

\subsection{Energy Score (Multivariate Proper Scoring Rule)}
\label{app:energy_score}

To assess the full multivariate predictive distribution in a distribution-agnostic manner, we compute the Energy Score (ES), a strictly proper scoring rule for multivariate forecasts. For a predictive random variable $X_i\sim F_i$ and an independent copy $X_i'$, the Energy Score is
\begin{equation}
\mathrm{ES}_i
=\mathbb{E}\left[\|X_i-y_i\|_2\right]
-\frac{1}{2}\mathbb{E}\left[\|X_i-X_i'\|_2\right],
\end{equation}
where $F_i$ is the method's predictive distribution (Gaussian for the ensemble, Student-$t$ for ours). Lower ES indicates better probabilistic predictions. We estimate the expectations via Monte Carlo with $S$ samples:
\begin{align}
\widehat{\mathrm{ES}}_i
&=\frac{1}{S}\sum_{s=1}^{S}\|x_i^{(s)}-y_i\|_2
-\frac{1}{2}\cdot\frac{1}{S}\sum_{s=1}^{S}\|x_i^{(s)}-\tilde{x}_i^{(s)}\|_2,\\
x_i^{(s)} &\overset{\text{i.i.d.}}{\sim} F_i,\qquad
\tilde{x}_i^{(s)} \overset{\text{i.i.d.}}{\sim} F_i.
\end{align}
We report the mean Energy Score over the evaluation set.

\section{Ensemble Predictive Covariance and $\sigma^2 I$ Calibration}
\label{app:ensemble_sigma_cal}

For a deterministic ensemble with member predictions $\{\mu_i^{(k)}\}_{k=1}^{K}$, we form the ensemble mean
\begin{equation}
\mu_i \;=\; \frac{1}{K}\sum_{k=1}^{K}\mu_i^{(k)},
\end{equation}
and an epistemic covariance estimate
\begin{equation}
\Sigma^{\text{epi}}_i
\;=\;
\frac{1}{K}\sum_{k=1}^{K}\left(\mu_i^{(k)}-\mu_i\right)\left(\mu_i^{(k)}-\mu_i\right)^\top.
\end{equation}
To improve calibration, we optionally add an isotropic variance floor
\begin{equation}
\Sigma_i(\sigma^2) \;=\; \Sigma^{\text{epi}}_i + \sigma^2 I,
\end{equation}
where $\sigma^2 \ge 0$ is fitted on a validation set by matching the observed coverage at a target level $p_{\mathrm{target}}$ (we use $p_{\mathrm{target}}=0.9$). Concretely, we choose $\sigma^2$ such that
\begin{equation}
\mathrm{Obs}_{\mathrm{val}}^{\sigma^2}(p_{\mathrm{target}}) \approx p_{\mathrm{target}},
\end{equation}
and then evaluate all metrics on the test set using the fixed $\sigma^2$.
\section{Equivariance Validation via Random Rotations}
\label{app:equivariance_validation}

In this section, we provide a numerical validation of the rotational
equivariance properties of the proposed model.
Specifically, we verify that the predicted forces and uncertainty tensors
transform correctly under arbitrary three-dimensional rotations.

\subsection{Experimental setup}

We randomly select a representative DWCT structure and apply a set of
independent random rotations.
For each trial, a rotation matrix $R \in \mathrm{SO}(3)$ is sampled uniformly,
and the atomic positions are transformed as
$\mathbf{x} \mapsto R \mathbf{x}$.

Let $\mathbf{F}(\mathbf{x}) \in \mathbb{R}^{N \times 3}$ denote the predicted
forces and let $\boldsymbol{U_{epi}}(\mathbf{x}) \in \mathbb{R}^{3 \times 3}$
denote the predicted uncertainty tensor.
For each rotation, we compute the deviations
\begin{equation}
\Delta \mathbf{F}
= \mathbf{F}(R\mathbf{x}) - R \mathbf{F}(\mathbf{x}),
\end{equation}
and
\begin{equation}
\Delta \boldsymbol{U_{epi}}
= \boldsymbol{U_{epi}}(R\mathbf{x}) - R \boldsymbol{U_{epi}}(\mathbf{x}) R^\top.
\end{equation}

The experiment is repeated for 300 randomly sampled rotations.
All components of the resulting deviation tensors are collected and flattened
to form empirical distributions.

\subsection{Results}

\begin{figure}
    \centering
    \includegraphics[width=\linewidth]{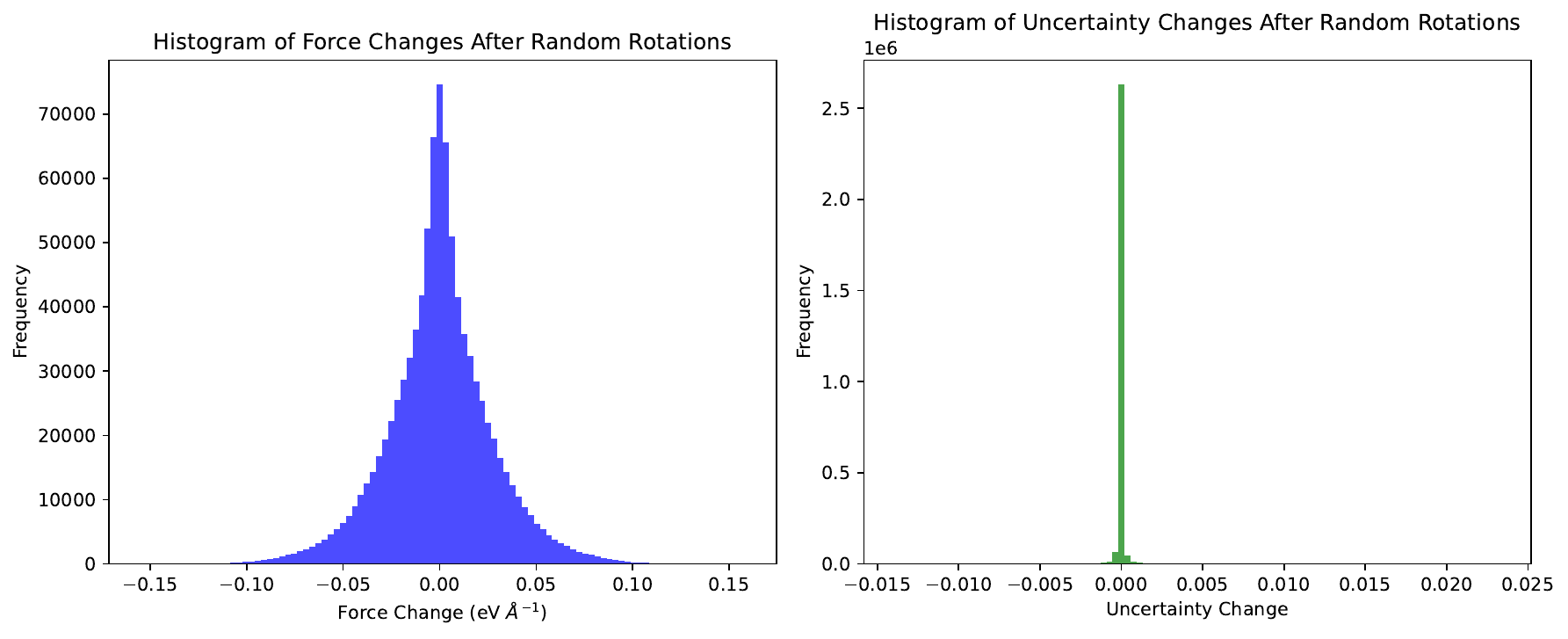}
    \caption{
    Histograms of deviations under random $\mathrm{SO}(3)$ rotations for a
    representative DWCT structure.
    \textbf{Left:} force deviations
    $\mathbf{F}(R\mathbf{x}) - R \mathbf{F}(\mathbf{x})$.
    \textbf{Right:} uncertainty deviations
    $\boldsymbol{\Sigma}(R\mathbf{x}) - R \boldsymbol{\Sigma}(\mathbf{x}) R^\top$.
    Both distributions are sharply peaked at zero, indicating that the model
    outputs satisfy rotational equivariance up to numerical precision.
    }
    \label{fig:rotation_equivariance_hist}
\end{figure}

Figure~\ref{fig:rotation_equivariance_hist} shows histograms of the force and
uncertainty deviations.
Both distributions are sharply centered around zero.

For the force predictions, the deviations exhibit a symmetric distribution
with small variance, indicating that the vector-valued outputs satisfy
rotational equivariance to numerical precision.
For the uncertainty predictions, the deviations are even more tightly
concentrated around zero, reflecting the stable equivariant behavior of the
second-order tensor output.

The observed residual deviations are several orders of magnitude smaller than
the typical scale of the model outputs and are attributed to floating-point
rounding errors and accumulated numerical precision limits.
These results confirm that the proposed model preserves $\mathrm{SO}(3)$
equivariance for both force and uncertainty predictions in practice.


\end{document}